\documentclass[lettersize,journal]{IEEEtran}
\usepackage{amsmath,amsfonts}
\usepackage{array}
\usepackage[caption=false,font=normalsize,labelfont=sf,textfont=sf]{subfig}
\usepackage{textcomp}
\usepackage{stfloats}
\usepackage{url}
\usepackage{verbatim}
\usepackage{graphicx}
\usepackage{cite}
\usepackage{amsmath}
\usepackage{mathrsfs}
\usepackage{amsopn}
\usepackage{cite}
\usepackage{amsthm}
\usepackage{epsfig,psfrag}
\usepackage{amssymb}
\usepackage{graphics}
\usepackage{epsfig}
\usepackage{epstopdf}
\usepackage{amsbsy}
\usepackage{amssymb} 
\usepackage{xcolor}
\usepackage{array}
\usepackage{arydshln}
\usepackage{color}
\usepackage{xcolor}
\usepackage{tabularx}
\usepackage{multirow}
\usepackage{xfrac}
\usepackage{subfig}
\usepackage{mathtools}
\usepackage{booktabs}
\usepackage{tabularx}
\usepackage{xcolor}
\usepackage{lipsum}
\usepackage{algorithm,algorithmic}
\usepackage{enumitem}
\usepackage{cleveref}
\usepackage{url} 
\usepackage{subfig}

\usepackage{amsfonts}       

\newcommand{\bb}{\boldsymbol}
\begin{document}

\title{An Efficient Approximate Method for Online Convolutional Dictionary Learning}

\author{Farshad G. Veshki and Sergiy A. Vorobyov,~\IEEEmembership{Fellow,~IEEE}
\thanks{The authors are with the Department of Information and Communications Engineering, Aalto University, Espoo, Finland (e-mail: farshad.ghorbaniveshki@aalto.fi; sergiy.vorobyov@aalto.fi)}
}


\maketitle

\begin{abstract}
Most existing convolutional dictionary learning (CDL) algorithms are based on batch learning, where the dictionary filters and the convolutional sparse representations are optimized in an alternating manner using a training dataset. When large training datasets are used, batch CDL algorithms become prohibitively memory-intensive. An online-learning technique is used to reduce the memory requirements of CDL by optimizing the dictionary incrementally after finding the sparse representations of each training sample. Nevertheless, learning large dictionaries using the existing online CDL (OCDL) algorithms remains highly computationally expensive. In this paper, we present a novel approximate OCDL method that incorporates sparse decomposition of the training samples. The resulting optimization problems are addressed using the alternating direction method of multipliers. Extensive experimental evaluations using several image datasets show that the proposed method substantially reduces computational costs while preserving the effectiveness of the state-of-the-art OCDL algorithms.  
 
\end{abstract}


\section{Introduction}
\IEEEPARstart{S}{parse} representations have become increasingly prevalent as a result of their wide use in diverse applications such as signal and image processing, machine learning, and computer vision~\cite{face_rec2009,im_sr2010,sig_rec2007,hyper2011}. The sparse representation model approximates a signal using a product of a matrix called a dictionary and a vector that only has a few non-zero entries (sparse representation). There are numerous applications where the use of the sparse representation model coupled with a learned dictionary results in remarkably improved performance. A learned dictionary aims to produce sparser representations and more accurate approximations of its domain signals~\cite{MOD1999,KSVD2006,ODL2009}.

Typically, dictionary learning and sparse approximation are used to extract local patterns and features from high-dimensional signals (such as images). Therefore, a prior decomposition of the original signals into vectorized overlapping blocks is usually required (e.g., patch extraction in image processing). However, relations between neighboring blocks are ignored, which results in multi-valued sparse representations and dictionaries composed of similar (shifted) atoms.

Convolutional sparse coding (CSC) provides a single-valued and shift-invariant model for describing high-dimensional signals~\cite{ShiftInvar1998,ShiftInvar2008,Papyan2017,Wohlberg2016}. This model replaces the matrix-vector product used in the standard sparse approximation by a sum of convolutions of dictionary filters $\{\bb{d}_k\in \mathbb{R}^{m}\}_{k=1}^K$ and convolutional sparse representations (CSRs) $\{\bb{x}_k\in \mathbb{R}^{P}\}_{k=1}^K$ (also called sparse feature maps). The convolutional sparse approximation problem can be formulated as follows
\begin{equation}
\begin{aligned}
\underset{\{\bb{x}_k\}_{k=1}^K}{\mathrm{minimize}} \; & \frac{1}{2}\left\|\sum_{k=1}^K \bb{d}_k \ast \bb{x}_k -\bb{s}\right\|_2^2+ \lambda\sum_{k=1}^K\|\bb{x}_k\|_1, \label{eq: CSC}
\end{aligned}
\end{equation}
where $\bb{s}\in \mathbb{R}^{P}$ is the signal, $\lambda>0$ is the regularization parameter that controls the sparsity of the representations, $\ast$ denotes the convolution operator (here, with ``$\textit{same}$'' padding), and $\|\cdot\|_1$ and $\|\cdot\|_2$ represent the $\ell_1$-norm and the Euclidean norm of a vector, respectively. 

The convolutional dictionary learning (CDL) problem is typically addressed using a batch approach in which the sparse representations and the dictionary filters are optimized alternately (batch CDL)~\cite{Wohlberg2016,Heide2015,Bristow2013,Choudhury2017,chalasani2013fast,Peng2019,veshki2021efficient}. The following is the formulation of the dictionary optimization problem over a batch of $N$ training signals $\{\bb{s}^n \in \mathbb{R}^{P}\}_{n=1}^N$,
\begin{equation}
\begin{split}{
\begin{aligned}
    \underset{\{\bb{d}_k\}_{k=1}^K}{\mathrm{minimize}} \; \frac{1}{2N}\sum_{n=1}^N\left\|\sum_{k=1}^K\bb{d}_k\ast\bb{x}^n_k-\bb{s}^n\right\|_2^2 + \sum_{k=1}^K\boldsymbol{\Omega}\left(\bb{d}_k\right), 
\label{eq: CDL}
\end{aligned} }
\end{split}
\end{equation}
where $\boldsymbol{\Omega}(\cdot)$ represents the indicator function of the constraint set for the dictionary filters, that is,
\begin{equation*}{
    \boldsymbol{\Omega}\left(\bb{d}\right) = 
     \begin{cases}
       0, &\quad\text{{\rm if}}\quad \|\bb{d}\|_2\leq1\\
       \infty, &\quad\text{\rm otherwise}.\\ 
     \end{cases}}
\end{equation*}

The existing batch CDL methods require access to all training signals and their CSRs at once. As a result, memory of the order of $NPK$ is required~\cite{cardona2018}, which can be extremely expensive when using large training datasets, i.e., when $N~\gg~K$. It is reminded that $K$ is the number of dictionary filters, $N$ is the number of training signals (the batch size), and $P$ is the dimension of the training signals, for example, the number of pixels in an image (usually $P\gg K$ and $P\gg N$). The memory requirement of CDL can be reduced using an online-learning approach, where the dictionary is optimized incrementally after observing each training signal and finding its sparse representations~\cite{ODL2009}. The online CDL (OCDL) methods are also useful when the training signals are not available all at once, but they are observed gradually over time. The state-of-the-art OCDL methods have achieved memory requirements of the order of $K^2P$~\cite{Wang2018,liu2018first}, which is independent of the number of training signals. Nevertheless, when learning large dictionaries or using high-dimensional signals, these methods can still incur excessive computational costs.

This paper presents a novel approximate OCDL method that significantly improves the computational efficiency of the state-of-the-art algorithms while providing competitive performance compared to the existing methods. As a result, we propose a method that requires a memory of the order of $KP$ only. More specifically, our method approximates the OCDL problem by minimizing an upper bound of the objective function, where the dictionary optimization problem is decentralized with respect to the convolutional filters. We then solve the resulting optimization problem using the \emph{alternating direction method of multipliers} (ADMM). MATLAB implementations of the proposed algorithms are available at \emph{\url{https://github.com/FarshadGVeshki/Approximate-Online-Convolutional-Dictionary-Learning}}.

The rest of the paper is organized as follows. Section~\ref{sec: background} briefly reviews CDL in the Fourier domain. The proposed CDL method and derivation of the algorithms are presented in detail in Section~\ref{sec: proposed}. Thorough experimental evaluation results in terms of convergence properties and reconstruction accuracy based on multiple image datasets of varying sizes are presented in Section~\ref{sec: results}. The conclusions are provided in Section~\ref{sec: conclusions}.

\section{OCDL in the Fourier Domain}
\label{sec: background}

Most efficient CDL methods are based on the Fourier transform~\cite{Wohlberg2016, veshki2021efficient,Wang2018,liu2018first}. In the frequency (Fourier) domain, problem \eqref{eq: CDL} is equivalent to
\begin{equation}
\begin{split}{
\begin{aligned}
    \underset{\{\bb{d}_k\}_{k=1}^K}{\mathrm{minimize}} \; \frac{1}{2NP}\sum_{n=1}^N\left\|\sum_{k=1}^K\hat{\bb{d}_k}\odot\hat{\bb{x}}^n_k-\hat{\bb{s}}^n\right\|_2^2 + \sum_{k=1}^K\boldsymbol{\Omega}\left(\bb{d}_k\right),
\label{eq: CDL Fourier}
\end{aligned} }
\end{split}
\end{equation}
where $(\hat{\cdot})$ and $\odot$ denote the discrete Fourier transform (DFT) and the elementwise multiplication operator, respectively. The filters $\{\bb{d}_k\}_{k=1}^K$ are zero-padded prior to DFT, so that $\{\hat{\bb{d}}_k\}_{k=1}^K$ are of the same size as the CSRs. 

Defining $\boldsymbol{\delta}_p \triangleq [ \hat{\boldsymbol{d}}_1(p), \cdots, \hat{\boldsymbol{d}}_K(p) ]^T$ and $ \boldsymbol{\chi}^n_p \triangleq [ \hat{\boldsymbol{x}}^n_1(p), \cdots, \hat{\boldsymbol{x}}^n_K(p)]^T$, problem~\eqref{eq: CDL Fourier} can be rewritten as
\begin{equation}
\begin{split}{
\begin{aligned}
    \underset{\{\bb{d}_k\}_{k=1}^K}{\mathrm{minimize}} \frac{1}{2NP}\sum_{p=1}^P\sum_{n=1}^N\left\|({\bb{\chi}^n_p})^T{\bb{\delta}}_p-\hat{\bb{s}}^n(p)\right\|_2^2 + \sum_{k=1}^K\boldsymbol{\Omega}\left(\bb{d}_k\right),
\label{eq: CDL Fourier re}
\end{aligned} }
\end{split}
\end{equation}
where $(\cdot)^T$ is the transpose operator. The most efficient solutions to problem \eqref{eq: CDL Fourier re} (the batch CDL problem) have been proposed based on ADMM, and the \emph{fast iterative shrinkage-thresholding algorithm} (FISTA)~\cite{veshki2021efficient,cardona2018}. The complexities of these algorithms are of $\mathcal{O}(KNP)$ and they require memory of the order of $KNP$. As a result, when the training dataset is large, batch CDL becomes excessively computationally demanding in practice. 

OCDL alleviates the problem of large required memory by storing sufficient statistics of the training signals and their CSRs in compact history arrays. An online reformulation of problem~\eqref{eq: CDL Fourier re} can be written as
\begin{equation}
\begin{split}{
\begin{aligned}
    \underset{\{\bb{d}_k\}_{k=1}^K}{\mathrm{minimize}} \; \frac{1}{2}\sum_{p=1}^P{\bb{\delta}}_p^H\bb{A}_p^N{\bb{\delta}}_p-\sum_{p=1}^P{\bb{\delta}}_p^T\bb{b}_p^N + \sum_{k=1}^K\boldsymbol{\Omega}\left(\bb{d}_k\right),
\label{eq: CDL Fourier re4}
\end{aligned} }
\end{split}
\end{equation}
where $(\cdot)^H$ is the Hermitian transpose operator, and the history arrays $\bb{A}^N_p \in \mathbb{R}^{K \times K}$ and $\bb{b}^N_p \in \mathbb{R}^{K}$, $p = 1,\dots,P$, are defined as
\begin{equation}
\begin{aligned}
\bb{A}^N_p \triangleq \frac{1}{NP}\sum_{n=1}^N(\bb{\chi}^n_p)^*({\bb{\chi}^n_p})^T,\quad
\bb{b}^N_p \triangleq \frac{1}{NP}\sum_{n=1}^N\hat{\bb{s}}^n(p)^* {\bb{\chi}^n_p},
\end{aligned}
\end{equation}
with $(\cdot)^*$ standing for the element-wise complex conjugate of an array vector. After observing each training signal and  finding its sparse representations, the history arrays are recalculated incrementally using the following formulas
\begin{equation}
\begin{aligned}
&\bb{A}^N_p = \frac{1}{NP} (\bb{\chi}^N_p)^*({\bb{\chi}^N_p})^T + \frac{N-1}{N}\bb{A}^{N-1}_p, \; p=1,\dots,P, \\
&\bb{b}^N_p = \frac{1}{NP}\hat{\bb{s}}^N(p)^* {\bb{\chi}^N_p} + \frac{N-1}{N}\bb{b}^{N-1}_p, \; p=1,\dots,P.
\end{aligned}
\end{equation}
The history arrays are initialized using zero arrays. In OCDL, the dictionary is optimized by solving problem~\eqref{eq: CDL Fourier re4} only after the updated history arrays are available. As a result, a memory requirement of $K^2P$ and a complexity of $\mathcal{O}(K^2NP)$ are achieved~\cite{Wang2018,liu2018first}.

\section{The Proposed Method}
\label{sec: proposed}

In the proposed method, the training signals are approximated in a distributed manner using $N$ distinct dictionaries $\{{\bb{c}}_k^n\in\mathbb{R}^{m}\}_{k=1}^K$. A fusion of the separately optimized dictionaries based on the respective CSRs is used to calculate the dictionary $\{{\bb{d}}_k\}_{k=1}^K$. Specifically, the quadratic term in CDL problem~\eqref{eq: CDL} is approximated using the following upper-bound estimate
\begin{equation}
\begin{aligned}
&\sum_{n=1}^N\left\|\sum_{k=1}^K\bb{d}_k\ast{\bb{x}}^n_k-\bb{s}^n\right\|_2^2\\
&\quad= \sum_{n=1}^N\left\|\sum_{k=1}^K{\bb{d}_k}\ast{\bb{x}}^n_k- \sum_{k=1}^K{\bb{c}}^n_k\ast{\bb{x}}^n_k+\sum_{k=1}^K{\bb{c}}^n_k\ast{\bb{x}}^n_k-{\bb{s}}^n\right\|_2^2 \\
&\quad\quad\leq \sum_{n=1}^N\sum_{k=1}^K\!\left\|{\bb{d}_k}\!\ast\!{\bb{x}}^n_k\!-\! {\bb{c}}^n_k\!\ast\!{\bb{x}}^n_k\right\|_2^2\!+\!\sum_{n=1}^N\left\|\sum_{k=1}^K\!{\bb{c}}^n_k\!\ast\!{\bb{x}}^n_k\!-\!{\bb{s}}^n\right\|_2^2,
\end{aligned}
\end{equation}
where the inequality is due to the triangle inequality. Accordingly, the proposed approximate CDL problem is formulated as
\begin{multline}
    \underset{\substack{\{\bb{d}_k\}_{k=1}^K,\\\{\{\bb{c}^n_k\}_{k=1}^K\}_{n=1}^N}}{\mathrm{minimize}} \frac{1}{2N}\!\sum_{n=1}^N\!\sum_{k=1}^K\!\left\|{\bb{d}_k}\!\ast\!{\bb{x}}^n_k\!-\!{\bb{c}}^n_k\!\ast\!{\bb{x}}^n_k\right\|_2^2 \!+\! \sum_{k=1}^K\!\boldsymbol{\Omega}\left(\bb{d}_k\right)\\
    +\! \frac{1}{2N}\!\sum_{n=1}^N\!\left\|\sum_{k=1}^K\!{\bb{c}}^n_k\!\ast\!{\bb{x}}^n_k\!-\!{\bb{s}}^n\right\|_2^2\!+\! \sum_{n=1}^N\! \sum_{k=1}^K\!\boldsymbol{\Omega}\left(\bb{c}^n_k\right).
\label{eq: CDL proposed}
\end{multline}

In the following, two ADMM-based online methods for addressing \eqref{eq: CDL proposed} are presented. The first algorithm uses a standard approach for optimization of $\{{\bb{d}}_k\}_{k=1}^K$ and $\{\bb{c}_k^N\}_{k=1}^K$, while the second algorithm incorporates pragmatic modifications to the first algorithm to improve the effectiveness of the proposed approximation method and lower computational costs.

\subsection{Algorithm 1}
\label{sec: alg1}
Optimization problem \eqref{eq: CDL proposed} is jointly convex with respect to $\{{\bb{d}}_k\}_{k=1}^K$ and $\{\{\bb{c}_k^n\}_{k=1}^K\}_{n=1}^N$. Thus, using the OCDL framework, problem \eqref{eq: CDL proposed} can be addressed for the joint optimization variables $\{\bb{c}_k^N,{\bb{d}}_k\}_{k=1}^K$ after observing the $N$th training signal $\bb{s}^N$ and obtaining its CSRs $\{\bb{x}_k^N\}_{k=1}^K$. Compact history arrays are used to store sufficient statistics of $\{\{\bb{c}_k^n\}_{k=1}^K\}_{n=1}^{N-1}$ and $\{\{\bb{x}_k^n\}_{k=1}^K\}_{n=1}^{N-1}$.

The following ADMM formulation is used to solve \eqref{eq: CDL proposed} for $\{\bb{c}_k^N,{\bb{d}}_k\}_{k=1}^K$
\label{sec: optimize D1}
\begin{equation}
 \begin{aligned}
    \underset{\substack{\{\bb{c}_k^N,\bb{d}_k\}_{k=1}^K,\\\{\bb{f}_k^N,\bb{g}_k\}_{k=1}^K}}{\mathrm{minimize}} \frac{1}{2N}\!\sum_{n=1}^N\!\sum_{k=1}^K\!\left\|{\bb{g}_k}\!\ast\!{\bb{x}}^n_k\!-\!{\bb{f}}^n_k\!\ast\!{\bb{x}}^n_k\right\|_2^2 \!+\! \sum_{k=1}^K\!\boldsymbol{\Omega}\left(\bb{d}_k\right)\\
    +\! \frac{1}{2N}\!\sum_{n=1}^N\!\left\|\sum_{k=1}^K\!{\bb{f}}^n_k\!\ast\!{\bb{x}}^n_k\!-\!{\bb{s}}^n\right\|_2^2\!+\! \sum_{n=1}^N\! \sum_{k=1}^K\!\boldsymbol{\Omega}\left(\bb{c}^n_k\right)\\
  \mathrm{s.t.}\quad\bb{g}_k = \bb{d}_k,\;\quad\bb{f}^N_k = \bb{c}^N_k,\; k=1,\dots,K,
\label{eq: OCDL D 1}
 \end{aligned}   
\end{equation}
where $\{\bb{f}_k^N,\bb{g}_k\}_{k=1}^K$ are the (joint) ADMM auxiliary variables. The ADMM iterations consist of the following three steps.
\subsubsection*{The $\{\bb{f},\bb{g}\}$-update step}
In this step the auxiliary variables $\{\bb{f}_k^N,\bb{g}_k\}_{k=1}^K$ are updated as
\begin{align}
\begin{split}
&\left(\{\bb{f}^N_k\}_{k=1}^K\right)^{t+1} = \underset{\{\bb{f}^N_k\}_{k=1}^K}{\mathrm{argmin}}  \frac{1}{2N}\sum_{k=1}^K\left\|{\bb{f}^N_k}\ast{\bb{x}}^N_k-{\bb{z}}^N_k\right\|_2^2\\
& +\! \frac{1}{2N}\!\left\|\sum_{k=1}^K{\bb{f}}^N_k\!\ast\!{\bb{x}}^N_k\!-\!{\bb{s}}^N\right\|_2^2 \!\!\!+\!\frac{\rho}{2}\sum_{k=1}^K \left\|{\bb{f}}^N_k\!-\! (\bb{c}^N_k)^{t} \!+\! ({\bb{u}}_k)^t\right\|_2^2,\label{eq: f-update}
\end{split}\\
\begin{split}
&\left(\{\bb{g}_k\}_{k=1}^K\right)^{t+1} = \underset{\{\bb{g}_k\}_{k=1}^K}{\mathrm{argmin}}  \frac{1}{2N}\sum_{n=1}^N\sum_{k=1}^K\left\|{\bb{g}_k}\ast{\bb{x}}^n_k-{\bb{t}}^n_k\right\|_2^2\\
&\hspace{5em}+\frac{\rho}{2}\sum_{k=1}^K \left\|{\bb{g}}_k- ({\bb{d}}_k)^{t} + ({\bb{v}}_k)^t\right\|_2^2,\label{eq: g-update}
\end{split}
\end{align}
where $\{\bb{u}_k,\bb{v}_k\}_{k=1}^K$ are the scaled Lagrangian variables, $\rho~>~0$ is the ADMM penalty parameter, $\bb{z}^N_k \triangleq (\bb{g}_k)^t\ast \bb{x}^N_k$ and $\bb{t}^n_k \triangleq (\bb{f}^n_k)^{t+1}\ast\bb{x}^n_k$.
\subsubsection*{The $\{\bb{c},\bb{d}\}$-update step}
In this step $\{\bb{c}_k^N,\bb{d}_k\}_{k=1}^K$ is updated as
\begin{align}
\begin{split}
&\big(\{\bb{c}^N_k\}_{k=1}^K\big)^{t+1} =  \underset{\{\bb{c}^N_k\}_{k=1}^K}{\mathrm{argmin}}  \; \sum_{k=1}^K\boldsymbol{\Omega}\left(\bb{c}^N_k\right)\\
&\hspace{7em}+\frac{\rho}{2}\sum_{k=1}^K \left\|(\bb{f}^N_k)^{t+1}- \bb{c}^N_k + (\bb{u}_k)^t\right\|_2^2,\label{eq: c-update}
\end{split}\\
\begin{split}
 &\big(\{\bb{d}_k\}_{k=1}^K\big)^{t+1} =  \underset{\{\bb{d}_k\}_{k=1}^K}{\mathrm{argmin}}  \; \sum_{k=1}^K\boldsymbol{\Omega}\left(\bb{d}_k\right)\\
&\hspace{7em}+\frac{\rho}{2}\sum_{k=1}^K \left\|(\bb{g}_k)^{t+1}- \bb{d}_k + (\bb{v}_k)^t\right\|_2^2.\label{eq: d-update}   
\end{split}
\end{align}

\subsubsection*{Updating the scaled Lagrangian parameters}
Finally, the scaled Lagrangian variables are updates as
\begin{align}
\begin{split}
&(\bb{u}_k)^{t+1} \!=\! (\bb{f}^N_k)^{t+1}\!-\!(\bb{c}^N_k)^{t+1}\!+\! (\bb{u}_k)^t,\quad k=1, \dots,K,\\
&(\bb{v}_k)^{t+1} \!=\! (\bb{g}_k)^{t+1}\!-\!(\bb{d}_k)^{t+1}\!+\! (\bb{v}_k)^t,\quad k=1, \dots,K.  \label{eq: scaled lagrangians}
\end{split}
\end{align}

The $\{\bb{c},\bb{d}\}$-update step involves projecting $(\bb{f}^N_k)^{t+1} + (\bb{u}_k)^t$ (in ~\eqref{eq: c-update}) and $(\bb{g}_k)^{t+1} + (\bb{v}_k)^t$ (in ~\eqref{eq: d-update}) onto the constraint set. First, the entries outside the support (${\mathbb{R}^{m}}$) are mapped to zero (recall that the filters are zero-padded), followed by projection onto the unit $\ell_2$-norm ball.

In the $\{\bb{f},\bb{g}\}$-update step, solving problem~\eqref{eq: f-update} is equivalent to solving the following optimization problem
\begin{multline}
    \underset{\{\bb{f}^N_k\}_{k=1}^K}{\mathrm{minimize}} \;\frac{1}{2N}\sum_{k=1}^K\left\|\hat{\bb{f}}^N_k\odot\hat{\bb{x}}^N_k-\hat{\bb{z}}^N_k\right\|_2^2 \\
    + \frac{1}{2N}\left\|\sum_{k=1}^K\hat{\bb{f}}^N_k\odot\hat{\bb{x}}^N_k-\hat{\bb{s}}^N\right\|_2^2 +\!\frac{\rho}{2}\!\sum_{k=1}^K \!\left\|\hat{\bb{f}}^N_k\!-\!\hat{\bb{q}}_k \right\|_2^2,
\label{eq: f update}
\end{multline}
where $\bb{q}_k \triangleq (\bb{c}^N_k)^{t} - ({\bb{u}}_k)^t$. By equating the derivative of the objective in~\eqref{eq: f update} to zero and using the Sherman-Morrison (SM) formula, the solution to the $\bb{f}$-update step is found as
\begin{equation}
\begin{aligned}
    \left(\hat{\boldsymbol{f}}^N_k(p)\right)^{t+1} &= \left( a^k_p + \frac{(a^k_p)^{2}|\hat{\bb{x}}_k^N\!(p)|^2}{1+\sum_{k=1}^K a^k_p|\hat{\bb{x}}_k^N(p)|^2}\right)\\
    &\times\left((\hat{\bb{x}}_k^N(p))^*\left(\hat{\bb{z}}_k^N(p) + \hat{\bb{s}}^N(p)\right) + N\rho \hat{\bb{q}}_k(p)\right),\label{eq: f-update sol}
\end{aligned}
\end{equation}
where $a^k_p\triangleq(|\hat{\bb{x}}_k^N(p)|^2+N\rho)^{-1}$. Using precalculated values of $\sum_{k=1}^K\! a^k_p|\hat{\bb{x}}_k^N\!(p)|^2$, the $\bb{f}$-update step can be carried out with the complexity of~$\mathcal{O}(KP)$ using~\eqref{eq: f-update sol}. 

Problem~\eqref{eq: g-update} can be addressed via solving the following optimization problem
\begin{equation}
\underset{\{\bb{g}_k\}_{k=1}^K}{\mathrm{minimize}}  \frac{1}{2N}\!\sum_{n=1}^N\!\sum_{k=1}^K\!\left\|\hat{\bb{g}_k}\!\odot\!\hat{\bb{x}}^n_k\!-\!\hat{\bb{t}}^n_k\right\|_2^2\!+\!\frac{\rho}{2}\!\sum_{k=1}^K \!\left\|\hat{\bb{g}}_k\!-\!\hat{\bb{w}}_k \right\|_2^2,\label{eq: d-update sol}
\end{equation}
where $\bb{w}_k \triangleq ({\bb{d}}_k)^{t} - ({\bb{v}}_k)^t$. 

The solution to~\eqref{eq: d-update sol} can be found as
\begin{equation}
    \left(\hat{\bb{g}}_k(p)\right)^{t+1} = \frac{\bb{\beta}_k^N(p)+\hat{\bb{w}}_k(p)}{\bb{\alpha}_k^N+\rho}, \; p=1,\dots,P,\;k=1,\dots,K,
\end{equation}
where history arrays $\bb{\alpha}^N_k \in \mathbb{R}^{P}$ and $\bb{\beta}^N_k \in \mathbb{R}^{P}$, $k \!= \!1,\dots,K$, are defined as
\begin{equation}
 \bb{\alpha}_k^N \triangleq \frac{1}{N}\sum_{n=1}^N  (\hat{\bb{x}}^n_k)^* \odot \hat{\bb{x}}^n_k, \;  \bb{\beta}_k^N \triangleq \frac{1}{N}\sum_{n=1}^N  (\hat{\bb{x}}^n_k)^* \odot \hat{\bb{t}}^n_k.
\end{equation}
The history arrays are incrementally updated using
\begin{equation}
 \bb{\alpha}_k^N = \frac{N-1}{N} \bb{\alpha}_k^{N-1}+ \frac{1}{N}(\hat{\bb{x}}^N_k)^* \odot \hat{\bb{x}}^N_k,
 \label{eq: update alpha}
\end{equation}
\begin{equation}
  \bb{\beta}_k^N = \frac{N-1}{N}\bb{\beta}_k^{N-1}+ \frac{1}{N}(\hat{\bb{x}}^N_k)^* \odot \hat{\bb{t}}^N_k.
  \label{eq: update beta}
\end{equation}

\Cref{alg: alg1} summarizes the main steps of the proposed approximate OCDL algorithm detailed in this section. Unit norm Gaussian distributed random arrays can be used as initial dictionary $\{\bb{d}^0_k\}_{k=1}^K$. At the first iteration, dictionary $\{\bb{d}_k\}_{k=1}^K$ can be used to initialize $\{\bb{c}^n_k\}_{k=1}^K$ and $\{\bb{g}_k\}_{k=1}^K$. Note that, before each iteration of the ADMM algorithm, $\{\bb{\beta}_k^n\}_{k=1}^K$ needs to be recalculated using~\eqref{eq: update beta} based on the latest values of $\{\bb{f}_k^n\}_{k=1}^K$.

 \begin{algorithm} 
 \caption{OCDL method proposed in Subsection~\ref{sec: alg1}}
 \begin{algorithmic}[1]\label{alg: alg1}
 \renewcommand{\algorithmicrequire}{\textbf{Input:}}
 \renewcommand{\algorithmicensure}{\textbf{Output:}}
 \REQUIRE Training signals $\{\bb{s}^n \in \mathbb{R}^{P}\}_{n=1}^N$, initial dictionary $\{\bb{d}^0_k\in \mathbb{R}^{m}\}_{k=1}^K$, sparsity regularization parameter $\lambda$;
 \\ \textit{Initialisation} : History arrays $\bb{\alpha}^0_k \in \mathbb{R}^{P}$ and $\bb{\beta}^0_k \in \mathbb{R}^{P}$, $k = 1,\dots,K$ as zero arrays, $\{\bb{d}_k\}_{k=1}^K=\{\bb{d}^0_k\}_{k=1}^K$;
 \FOR {$n = 1$ to $N$}
  \STATE Find $\{\bb{x}^n_k\}_{k=1}^K$ for $\bb{s}^n$ using $\{\bb{d}_k\}_{k=1}^K$ and $\lambda$ by solving \eqref{eq: CSC};
  \STATE Calculate $\{\bb{\alpha}^n_k\}_{k=1}^K$ using \eqref{eq: update alpha};
  \STATE Optimize $\{\bb{c}^n_k,\bb{d}_k\}_{k=1}^K$ using the ADMM-based method in Subsection~\ref{sec: alg1} (recalculate $\{\bb{\beta}_k^n\}_{k=1}^K$ using~\eqref{eq: update beta} in every iteration);
  \ENDFOR
 \RETURN Learned convolutional dictionary $\{\bb{d}_k\}_{k=1}^K$.
 \end{algorithmic} 
 \end{algorithm}

\subsection{Algorithm 2}
\label{sec: alg2}
To improve the performance of the proposed OCDL algorithm, dictionary optimization can be performed \emph{exactly} for the latest observed signal $\bb{s}^N$, while the proposed approximation method is used for $\{\bb{s}^{n}\}_{n=1}^{N-1}$. Thus, the modified approximate CDL problem is now formulated as
\begin{multline}
    \underset{\substack{\{\bb{d}_k\}_{k=1}^K,\\\{\{\bb{c}^n_k\}_{k=1}^K\}_{n=1}^N}}{\mathrm{minimize}} 
    \frac{1}{2N}\!\!\left\|\sum_{k=1}^K\!{\bb{d}}_k\!\ast\!{\bb{x}}^N_k\!-\!{\bb{s}}^N\right\|_2^2\!\\
    +\frac{1}{2N}\!\sum_{n=1}^{N-1}\!\sum_{k=1}^K\!\left\|{\bb{d}_k}\!\ast\!{\bb{x}}^n_k\!-\!{\bb{c}}^n_k\!\ast\!{\bb{x}}^n_k\right\|_2^2 \!+\! \sum_{k=1}^K\!\boldsymbol{\Omega}\left(\bb{d}_k\right)\\
    +\! \frac{1}{2N}\!\sum_{n=1}^{N-1}\!\left\|\sum_{k=1}^K\!{\bb{c}}^n_k\!\ast\!{\bb{x}}^n_k\!-\!{\bb{s}}^n\right\|_2^2\!+\! \sum_{n=1}^N\! \sum_{k=1}^K\!\boldsymbol{\Omega}\left(\bb{c}^n_k\right).
\label{eq: CDL proposed 2}
\end{multline}
The alternating procedure for addressing~\eqref{eq: CDL proposed 2} consists of the following steps.
\subsubsection{Optimization of $\{\bb{d}_k\}_{k=1}^K$}
\label{sec: optimize D2}
Solving~\eqref{eq: CDL proposed 2} with respect to $\{\bb{d}_k\}_{k=1}^K$ can be addressed using the following ADMM formulation
\begin{multline}
    \underset{\{\bb{d}_k\}_{k=1}^K,\{\bb{g}_k\}_{k=1}^K}{\mathrm{minimize}} 
    \frac{1}{2N}\left\|\sum_{k=1}^K{\bb{g}}_k\ast{\bb{x}}^N_k-{\bb{s}}^N\right\|_2^2\\
    +\frac{1}{2N}\sum_{n=1}^{N-1}\sum_{k=1}^K\left\|{\bb{g}_k}\ast{\bb{x}}^n_k-{\bb{r}}^n_k\right\|_2^2 + \sum_{k=1}^K\boldsymbol{\Omega}\left(\bb{d}_k\right)\\
   \mathrm{s.t.}\quad \bb{g}_k = \bb{d}_k,\; k=1,\dots,K.
\label{eq: OCDL D 2}
\end{multline}
where $\bb{r}^n_k \triangleq \bb{c}^n_k\ast\bb{x}^n_k$. 

The ADMM iterations consist of the following steps: 
\begin{itemize}
    \item[(i)] the $\bb{g}$-update step: a convolutional least-squares fitting problem);
    \item[(ii)] the $\bb{d}$-update step: projection on the constraint set (similar to~\eqref{eq: d-update});
    \item[(iii)] updating the Lagrangian multipliers (similar to \eqref{eq: scaled lagrangians}).
\end{itemize}

The $\bb{g}$-update step requires solving the optimization problem in the form of
\begin{multline}
\underset{\{\bb{g}_k\}_{k=1}^K}{\mathrm{minimize}} \; \frac{1}{2N}\left\|\sum_{k=1}^K\!\hat{\bb{g}_k}\odot\hat{\bb{x}}^N_k-\hat{\bb{s}}^N\right\|_2^2\\
+\frac{1}{2N}\!\sum_{n=1}^{N-1}\!\sum_{k=1}^K\!\left\|\hat{\bb{g}_k}\!\odot\!\hat{\bb{x}}^n_k\!-\!\hat{\bb{r}}^n_k\right\|_2^2\!
+\!\frac{\rho}{2}\!\sum_{k=1}^K \!\left\|\hat{\bb{g}}_k\!-\!\hat{\bb{e}}_k \right\|_2^2.\label{eq: d-update sol 2}
\end{multline}
Equating the derivative to zero and using the SM formula, optimization problem \eqref{eq: d-update sol 2} can be solved as
\begin{equation}
\begin{aligned}
    \left(\hat{\boldsymbol{g}}^N_k(p)\right)^{t+1} &= \left( b^k_p + \frac{(b^k_p)^{2}|\hat{\bb{x}}_k^N(p)|^2}{N
    +\sum_{k=1}^K b^k_p|\hat{\bb{x}}_k^N(p)|^2}\right)\\
   & \times\left( \frac{1}{N}(\hat{\bb{x}}_k^N(p))^*\hat{\bb{s}}^N(p) + \bb{\tilde{\beta}}_k^{N-1}(p) +\rho \hat{\bb{e}}_k(p)\right), \label{eq: g-update rule 2}
\end{aligned}
\end{equation}
with $b^k_p\triangleq(\bb{\tilde{\alpha}}_k^{N-1}(p)+\rho)^{-1}$, where history arrays $\bb{\tilde{\alpha}}^N_k \in \mathbb{R}^{P}$ and $\bb{\tilde{\beta}}^N_k \in \mathbb{R}^{P}$, $k \!= \!1,\dots,K$, are defined as
\begin{equation}
 \bb{\tilde{\alpha}}_k^N \triangleq \frac{1}{N+1}\sum_{n=1}^N  (\hat{\bb{x}}^n_k)^* \odot \hat{\bb{x}}^n_k, \;  \bb{\tilde{\beta}}_k^N \triangleq \frac{1}{N+1}\sum_{n=1}^N  (\hat{\bb{x}}^n_k)^* \odot \hat{\bb{r}}^n_k.
\end{equation}
The incremental update rules for $\bb{\tilde{\alpha}}_k^N$ and $\bb{\tilde{\beta}}_k^N$ can be found as
\begin{equation}
 \bb{\tilde{\alpha}}_k^N = \frac{N}{N+1} \bb{\tilde{\alpha}}_k^{N-1}+ \frac{1}{N+1}(\hat{\bb{x}}^N_k)^* \odot \hat{\bb{x}}^N_k,
 \label{eq: update alpha 2}
\end{equation}
\begin{equation}
  \bb{\tilde{\beta}}_k^N = \frac{N}{N+1}\bb{\tilde{\beta}}_k^{N-1}+ \frac{1}{N+1}(\hat{\bb{x}}^N_k)^* \odot \hat{\bb{r}}^N_k.
  \label{eq: update beta 2}
\end{equation}

The $\bb{g}$-update~\eqref{eq: g-update rule 2} can be performed with the complexity of~$\mathcal{O}(KP)$ using precalculated values of $\sum_{k=1}^K\! b^k_p|\hat{\bb{x}}_k^N\!(p)|^2$.

\subsubsection{Optimization of $\{\bb{c}_k^N\}_{k=1}^K$}
\label{sec: optimize C2}
In the modified algorithm, dictionary $\{\bb{c}_k^N\}_{k=1}^K$ is optimized only to provide a more accurate approximation of $\bb{s}^N$ (in comparison with the approximation provided using $\{\bb{d}_k\}_{k=1}^K$). It means that the second quadratic term in \eqref{eq: CDL proposed 2} is ignored in the step of $\{\bb{c}_k^N\}_{k=1}^K$ optimization. Here we rely on the fact that CSRs $\{{\bb{x}}^N_k\}_{k=1}^K$ are direct products of $\{\bb{d}_k\}_{k=1}^K$. As a result, considering that the approximation is based on $\{{\bb{x}}^N_k\}_{k=1}^K$, the resulting $\{\bb{c}_k^N\}_{k=1}^K$ cannot unfavorably deviate from $\{\bb{d}_k\}_{k=1}^K$. Problem~\eqref{eq: CDL proposed 2}, which needs to be solved now for $\{\bb{c}_k^N\}_{k=1}^K$ only, is then reduced to the following optimization problem
\begin{equation}
    \underset{\{\bb{c}^N_k\}_{k=1}^K}{\mathrm{minimize}} 
    \frac{1}{2P}\left\|\sum_{k=1}^K{\bb{c}}^N_k\ast{\bb{x}}^N_k-{\bb{s}}^N\right\|_2^2+\sum_{k=1}^K\boldsymbol{\Omega}\left(\bb{c}^N_k\right),
\label{eq: OCDL C 2}
\end{equation}
which is a CDL problem involving a single training signal, which can be addressed using the existing CDL methods (e.g., \cite{veshki2021efficient}).

The main steps of the presented approximate OCDL algorithm are summarized in \Cref{alg: alg2}. Optimization of dictionaries $\{\bb{d}_k\}_{k=1}^K$ and $\{\bb{c}_k^n\}_{k=1}^K$ (lines $3$ and $4$) can be initialized using the existing $\{\bb{d}_k\}_{k=1}^K$.

 \begin{algorithm} 
 \caption{OCDL method proposed in Subsection~\ref{sec: alg2}}
 \begin{algorithmic}[1]\label{alg: alg2}
 \renewcommand{\algorithmicrequire}{\textbf{Input:}}
 \renewcommand{\algorithmicensure}{\textbf{Output:}}
 \REQUIRE Training signals $\{\bb{s}^n \in \mathbb{R}^{P}\}_{n=1}^N$, initial dictionary $\{\bb{d}^0_k\in \mathbb{R}^{m}\}_{k=1}^K$, sparsity regularization parameter $\lambda$;
 \\ \textit{Initialisation} : History arrays $\bb{\tilde{\alpha}}^0_k \in \mathbb{R}^{P}$ and $\bb{\tilde{\beta}}^0_k \in \mathbb{R}^{P}$, $k = 1,\dots,K$ as zero arrays, $\{\bb{d}_k\}_{k=1}^K=\{\bb{d}^0_k\}_{k=1}^K$;
 \FOR {$n = 1$ to $N$}
  \STATE Find $\{\bb{x}^n_k\}_{k=1}^K$ for $\bb{s}^n$ and $\{\bb{d}_k\}_{k=1}^K$ by solving \eqref{eq: CSC};
  \STATE Optimize $\{\bb{d}_k\}_{k=1}^K$ as in Subsection~\ref{sec: optimize D2};
  \STATE Optimize $\{\bb{c}^n_k\}_{k=1}^K$ as in Subsection~\ref{sec: optimize C2};
  \STATE Calculate $\{\bb{\tilde{\alpha}}^n_k\}_{k=1}^K$ and $\{\bb{\tilde{\beta}}^n_k\}_{k=1}^K$ using \eqref{eq: update alpha 2}, \eqref{eq: update beta 2};
  \ENDFOR
 \RETURN learned convolutional dictionary $\{\bb{d}_k\}_{k=1}^K$.
 \end{algorithmic} 
 \end{algorithm}

\subsection{Memory Requirements and Computational Complexity}
The largest arrays used in the proposed algorithms are of size $KP$. The most computationally expensive steps of performing updates \eqref{eq: f-update sol} and \eqref{eq: g-update rule 2} both have a complexity of $\mathcal{O}(KP)$, which is slightly dominated by the complexity of DFT that is of $\mathcal{O}(KP\mathrm{log}(P))$ when performed using \emph{Fast Fourier Transform}. Thus, the computational complexity of the proposed algorithm is of the order of $KP$ sequentially performed $N$ times (once for each signal in the training dataset).

\section{Experimental Results}
\label{sec: results}
\subsection{Compared Methods}
The performance of the proposed algorithms is benchmarked against the following state-of-the-art OCDL methods:
\begin{description}
    \item[\textbf{OCSC}] The ADMM-based OCDL method of~\cite{Wang2018}, which uses the iterative Sherman-Morrison formula for updating the history arrays; 
    \item[\textbf{FISTA}] The FISTA-based OCDL method of\cite{liu2018first} that uses gradient calculated in the Fourier domain. 
\end{description}
In addition, we compare the OCDL methods to the following batch-CDL algorithm,
\begin{description}
    \item[\textbf{ADMM-cns}] The batch-CDL method of~\cite{veshki2021efficient} that is based on consensus-ADMM. 
\end{description}

Algorithms~\ref{alg: alg1}-\ref{alg: alg2} are referred to as ``proposed-1'' and ``proposed-2'', respectively.

\subsection{Datasets}
The experiments are conducted using the following $5$ image datasets:
\begin{description}
    \item[Fruit] and \textbf{City} Two small datasets, each composed of 10 images of size $100\times 100$. These datasets are typically used as benchmarks for CSC and CDL~\cite{Bristow2013,Heide2015,Wang2018};
    \item[SIPI] A dataset composed of $20$ training images and $5$ test images all of size $256\times 256$ collected from the UCS-SIPI image database~{\em \url{http://sipi.usc.edu/database/}}.
    \item[Flicker] A dataset composed of $40$ training images and $5$ test images all of size $256\times 256$ collected from the MIRFLICKR-1M image dataset~{\em \url{https://press.liacs.nl/mirflickr/mirdownload.html}}.
    \item[Flicker-large] A dataset composed of $1000$ training images and $50$ test images all of size $256\times 256$ collected from the MIRFLICKR-1M image dataset.
\end{description}

The initial images are transformed into greyscale and the $8$-bit pixel values are normalized to a range of 0-1 by dividing by 255. Images from the MIRFLICKR-1M and USC-SIPI datasets are then cropped and resized. As the CSC model is not capable of effectively handling low-frequency signals, it is a common practice to use high-pass filtered images for CDL~\cite{cardona2018,Wohlberg2016,liu2018first}. In the experiments, the low-frequency components of all images are eliminated using the \textit{lowpass} function of the SPORCO toolbox~\cite{wohlberg-2016-sporco} with a regularization parameter of $5$.

\begin{figure*}[t]
\centering{
\includegraphics[width=.99\linewidth]{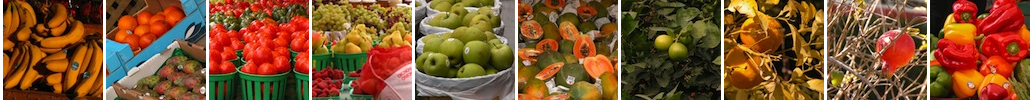}\\\vspace{0.5mm}
\includegraphics[width=.99\linewidth]{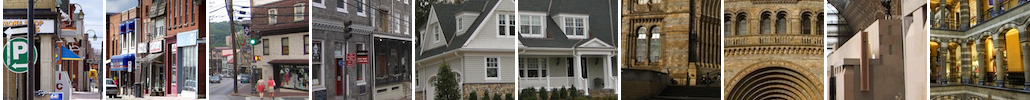}}
\caption{Datsets \emph{Fruit} (first row) and \emph{City} (second row).}
\label{fig: small data}
\end{figure*}

\subsection{Implementation Details}
The proposed algorithms employ the unconstrained convolutional sparse approximation method of~\cite{veshki2021efficient}. In all ADMM-based algorithms (both sparse approximation and dictionary learning) the maximum number of iterations is set to $300$, and stopping criteria discussed in~\cite[Subsection~3.3]{boyd2011distributed} with absolute and relative tolerance values of $10^{-4}$ are used. We use dictionary filters of size $8\times 8$ in all experiments.

All ADMM-based algorithms except OCSC use ADMM extensions \textit{over-relaxation}~\cite[Subsection~3.4.3]{boyd2011distributed} and \textit{varying penalty parameter}~\cite[Subsection~3.4.1]{boyd2011distributed} with initial penalty parameter $\rho=10$ (the same parameters are used in all methods). The OCSC method incorporates the ADMM penalty parameter $\rho$ in the history arrays. Thus, this method cannot use \textit{varying penalty parameter} extension. For the OCSC method, we use the default parameters set by the authors of the paper (the stopping criteria are modified to be uniform with other algorithms compared).

In all experiments, we use $\lambda=0.1\lambda_{\rm max}$, where $\lambda_{\rm max}$ is the smallest value that results in all-zero sparse representations and can be obtained using $\ell_{\infty}$-norm of the gradient of the objective of convolutional sparse approximation problem~\eqref{eq: CSC} at $\{\bb{x}_k\}_{k=1}^K=\bb{0}$. Here, the value of $\lambda_{\rm max}$ is calculated only once using the first image in the training datasets.

All algorithms are implemented using MATLAB. All experiments are performed using a PC equipped with an Intel(R) Core(TM) i5-8365U 1.60GHz CPU and 16GB memory.

\subsection{Comparison Criteria}
The effectiveness of the CDL algorithms is typically evaluated based on the objective values of the convolutional sparse approximation problem~\eqref{eq: CSC} averaged over the entire test datasets~\cite{liu2018first,Wang2018,Zisselman2019}. A lower objective value indicates a better performance. For the small datasets \emph{Fruit} and \emph{City}, since there is no test data, the average training objective values are reported to compare the effectiveness of the optimization algorithms~\cite{Heide2015}. Using visualized learned dictionary filters, the OCDL algorithms are evaluated for their ability to extract (learn) visual features. The efficiency of the algorithms is measured using the training times. 

\subsection{Small Datasets Fruit and City}
\begin{figure}[htb]
\centering
{\includegraphics[width=.65\linewidth]{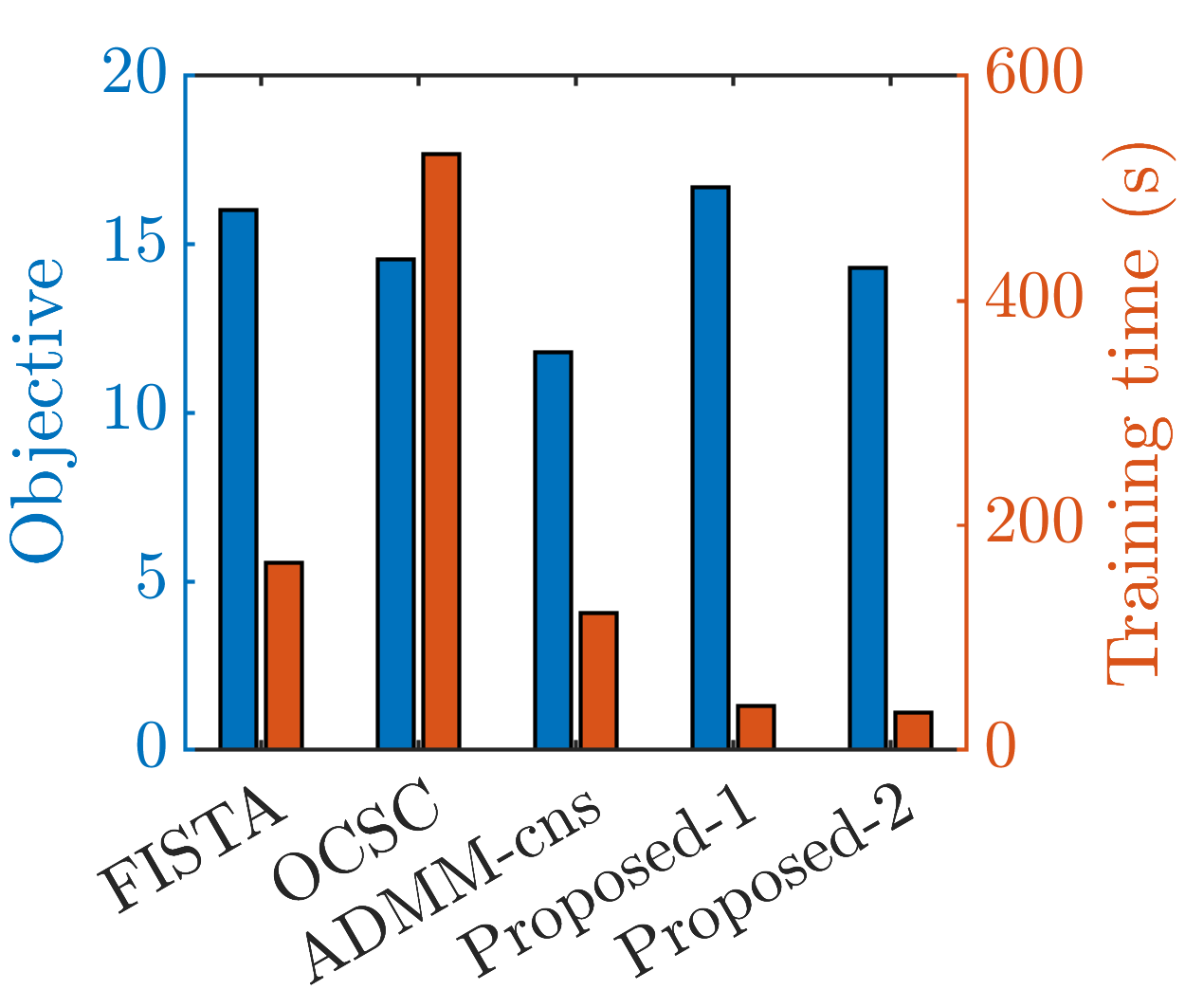}}\\
{\includegraphics[width=.65\linewidth]{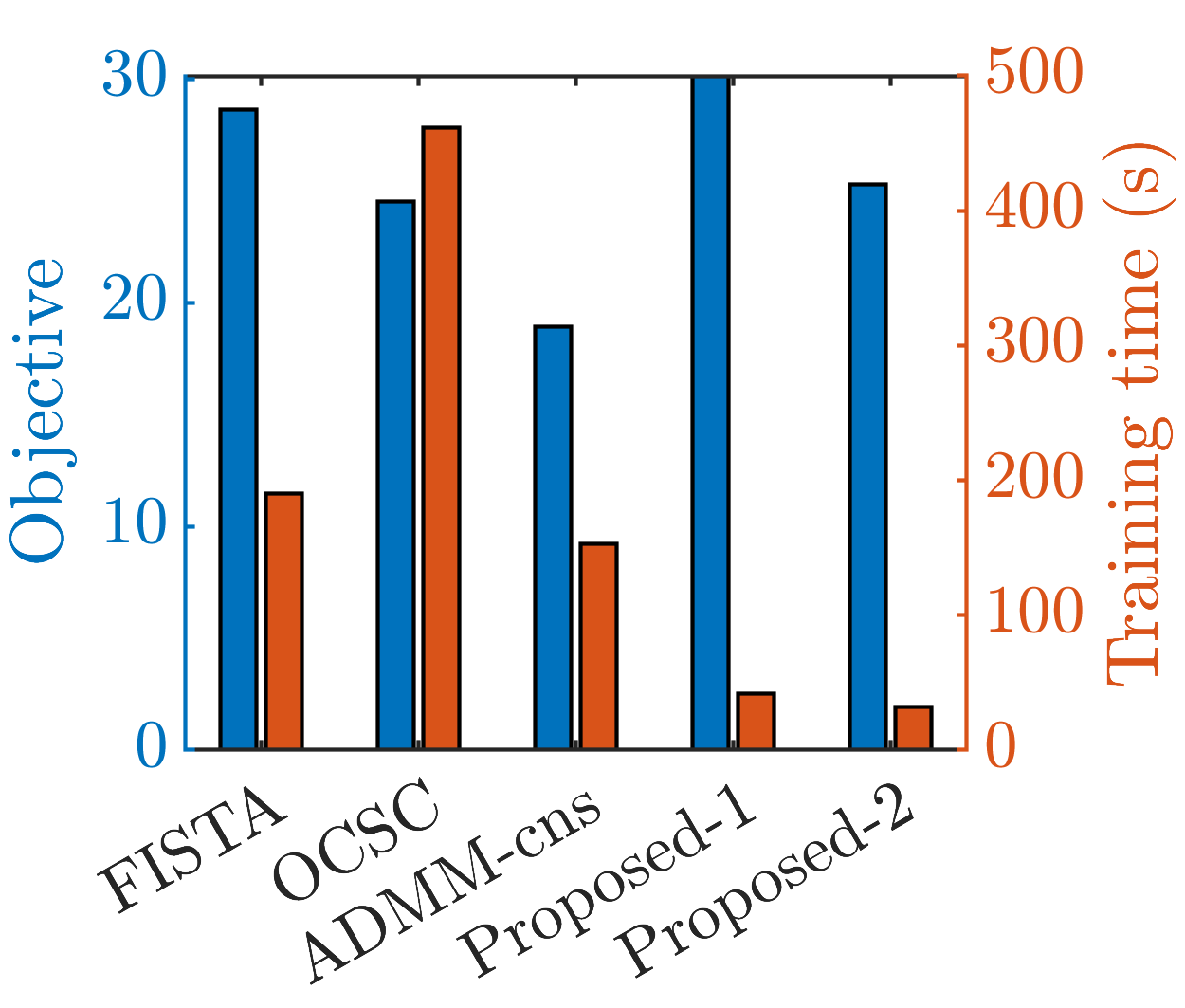}}
\caption{Comparison of training objective values and training times obtained using all methods compared for datasets \emph{Fruit} (top) and \emph{City} (bottom).}
\label{fig: bars small}
\end{figure}

Fig.~\ref{fig: small data} shows the images in the small datasets \emph{Fruit} and \emph{City}. \Cref{tab: city,tab: fruit} report the average training objective values and the training times obtained using the methods tested for these two datasets. To facilitate comparison, the results are presented as bar plots in Fig.~\ref{fig: bars small}. The experiments based on datasets \emph{Fruit} and \emph{City} are performed using dictionary size $K=64$.

\begin{table}[htb]
\caption{Average training objective values and training times obtained using the methods compared for dataset \emph{Fruit}.}
\centering
\setlength{\tabcolsep}{10pt}
\renewcommand{\arraystretch}{1}
\begin{tabular}{|c|c|c|}
\hline & Objective & Training Time (s) \\ \hline
Initial dictionary                     & $19.5422$   & -                 \\ \hline
FISTA~\cite{liu2018first}              & $16.0159$   & $167$          \\ \hline
OCSC~\cite{Wang2018}                   & $14.5529$   & $530$          \\ \hline
Proposed-1                             & $16.6867$   & $39$           \\ \hline
Proposed-2                             & $14.3059$   & $33$           \\ \hline
ADMM-cns (batch)~\cite{veshki2021efficient} & $11.8088$   & $122$          \\ \hline
\end{tabular}\label{tab: fruit}
\end{table}

\begin{table}[htb]
\caption{Average training objective values and training times obtained using the methods compared for dataset \emph{City}.}
\centering
\setlength{\tabcolsep}{10pt}
\renewcommand{\arraystretch}{1}
\begin{tabular}{|c|c|c|}
\hline & Objective & Training Time (s) \\ \hline 
Initial dictionary               & $33.9411$          & -      \\ \hline
FISTA~\cite{liu2018first}        & $28.6235$          & $190$                \\ \hline
OCSC~\cite{Wang2018}             & $24.5472$          & $462$             \\ \hline
Proposed-1                       & $30.1463$          & $42$                 \\ \hline
Proposed-2                       & $25.2740$          & $32$                  \\ \hline
ADMM-cns (batch)~\cite{veshki2021efficient} & $18.9411$     & $153$           \\ \hline
\end{tabular}\label{tab: city}
\end{table}

As can be observed, the ADMM-cns batch CDL algorithm yields the lowest objective function values. However, this method is not suitable for large datasets as mentioned earlier. The proposed methods produce objective values that are comparable to other OCDL algorithms tested. In particular, algorithm 2 (proposed-2) results in the smallest objective for the \emph{Fruit} dataset among all OCDL algorithms. For the \emph{City} dataset, the OCSC method has the lowest objective compared to other OCDL methods (slightly better than that of proposed-2), but shows a longer training time. As shown in \Cref{tab: city,tab: fruit}, the proposed algorithms result in substantially shorter training times, especially Algorithm~2, which is noticeably faster than Algorithm~1.

\begin{figure}[htb]
\centering{
\subfloat[ADMM-cns]{\includegraphics[width=.3\linewidth]{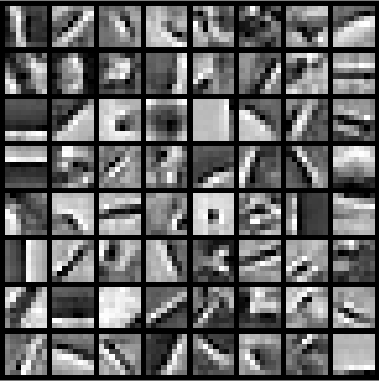}}\,
\subfloat[FISTA]{\includegraphics[width=.3\linewidth]{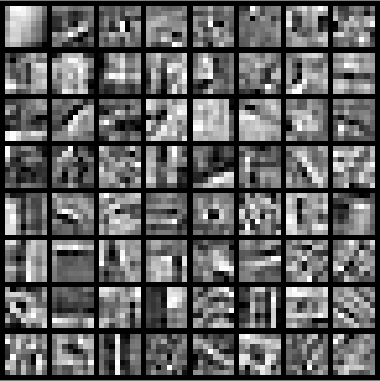}}\,
\subfloat[OCSC]{\includegraphics[width=.3\linewidth]{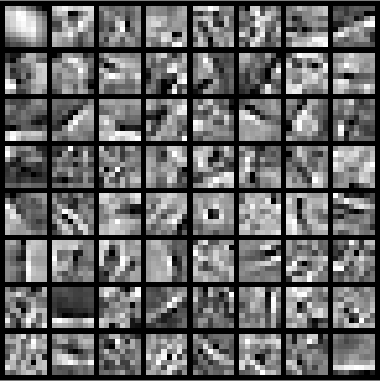}}\\\vspace{0.5mm}
\subfloat[Proposed-1]{\includegraphics[width=.3\linewidth]{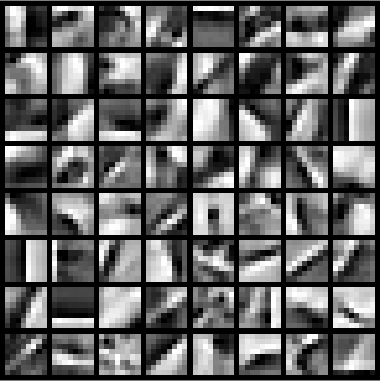}}\,
\subfloat[Proposed-2]{\includegraphics[width=.3\linewidth]{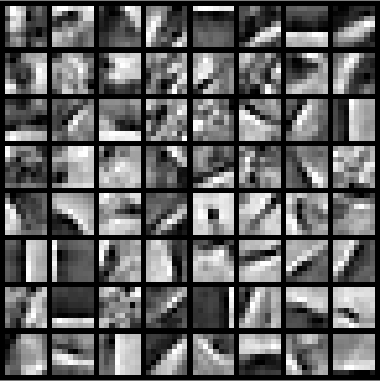}}}
\caption{Dictionaries learned ($K=64$) using the methods compared for dataset \emph{Fruit}.}
\label{fig: fruit dicts}
\end{figure}

\begin{figure}[htb]
\centering{
\subfloat[ADMM-cns]{\includegraphics[width=.3\linewidth]{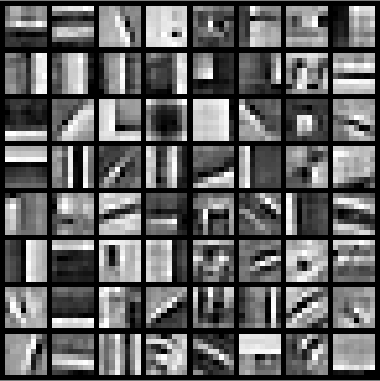}}\,
\subfloat[FISTA]{\includegraphics[width=.3\linewidth]{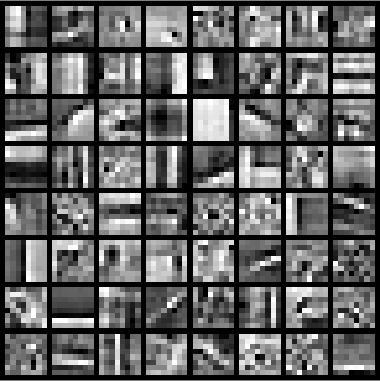}}\,
\subfloat[OCSC]{\includegraphics[width=.3\linewidth]{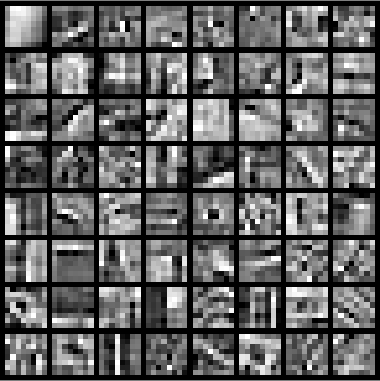}}\\\vspace{0.5mm}
\subfloat[Proposed-1]{\includegraphics[width=.3\linewidth]{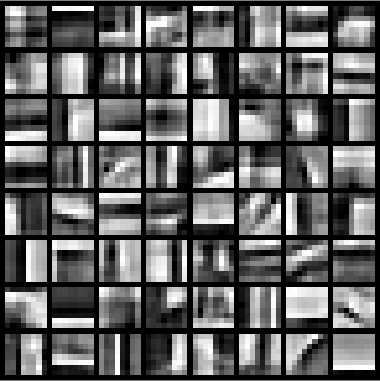}}\,
\subfloat[Proposed-2]{\includegraphics[width=.3\linewidth]{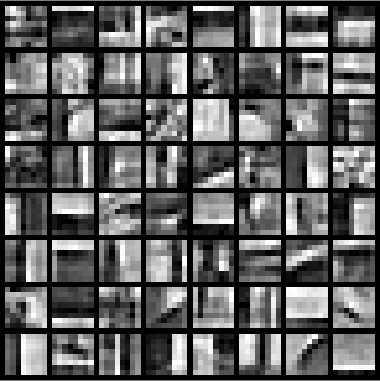}}}
\caption{Dictionaries learned ($K=64$) using the methods compared for dataset \emph{City}.}
\label{fig: city dicts}
\end{figure}

The convolutional dictionaries learned based on datasets \emph{Fruit} and \emph{City} using the methods tested are visualized in Figs.~\ref{fig: fruit dicts} and \ref{fig: city dicts}, respectively. 

Acquiring \emph{valid} (as opposed to \emph{noisy} and \emph{random}) visual features is crucial in many image and signal processing tasks that utilize dictionary learning, such as image denoising, image inpainting, and image fusion. By examining the dictionaries shown in Figs.~\ref{fig: fruit dicts} and \ref{fig: city dicts}, it can be seen that the dictionaries learned using the proposed method contain fewer noisy and random filters compared to those learned using OCSC and FISTA. The filters in the dictionaries learned using ADMM-cns (batch CDL) appear \emph{crisper} and \emph{sharper}, while those learned using the proposed algorithms seem \emph{smoother}. This can be explained by the fact that in the proposed method, the dictionaries are, in a way, learned from the sparse approximation of the original images.

\subsection{Datasets SIPI and Flickr}
\begin{figure*}[htb]
\centering
\includegraphics[width=.99\linewidth]{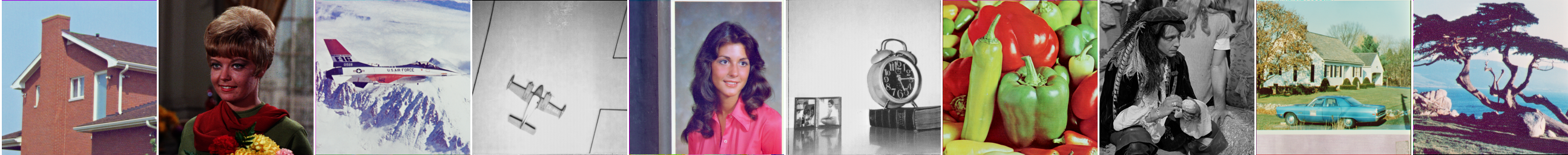}
\caption{$10$ randomly selected images from dataset \emph{SIPI}.}
\label{fig: SIPI data}
\end{figure*}

\begin{figure*}[htb]
\centering
\includegraphics[width=.99\linewidth]{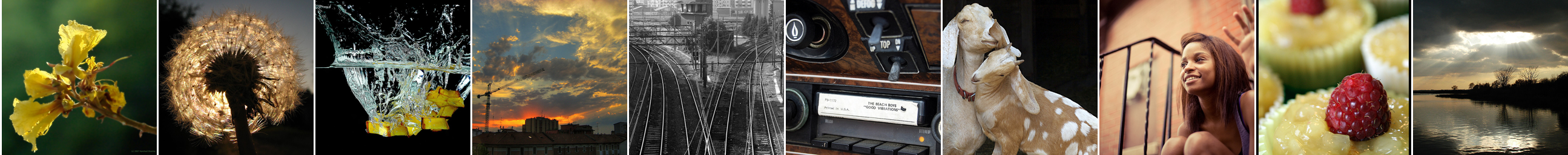}
\caption{$10$ randomly selected images from dataset \emph{Flickr}.}
\label{fig: Flickr data}
\end{figure*}

\begin{figure}[htb]
\centering
{\includegraphics[width=.65\linewidth]{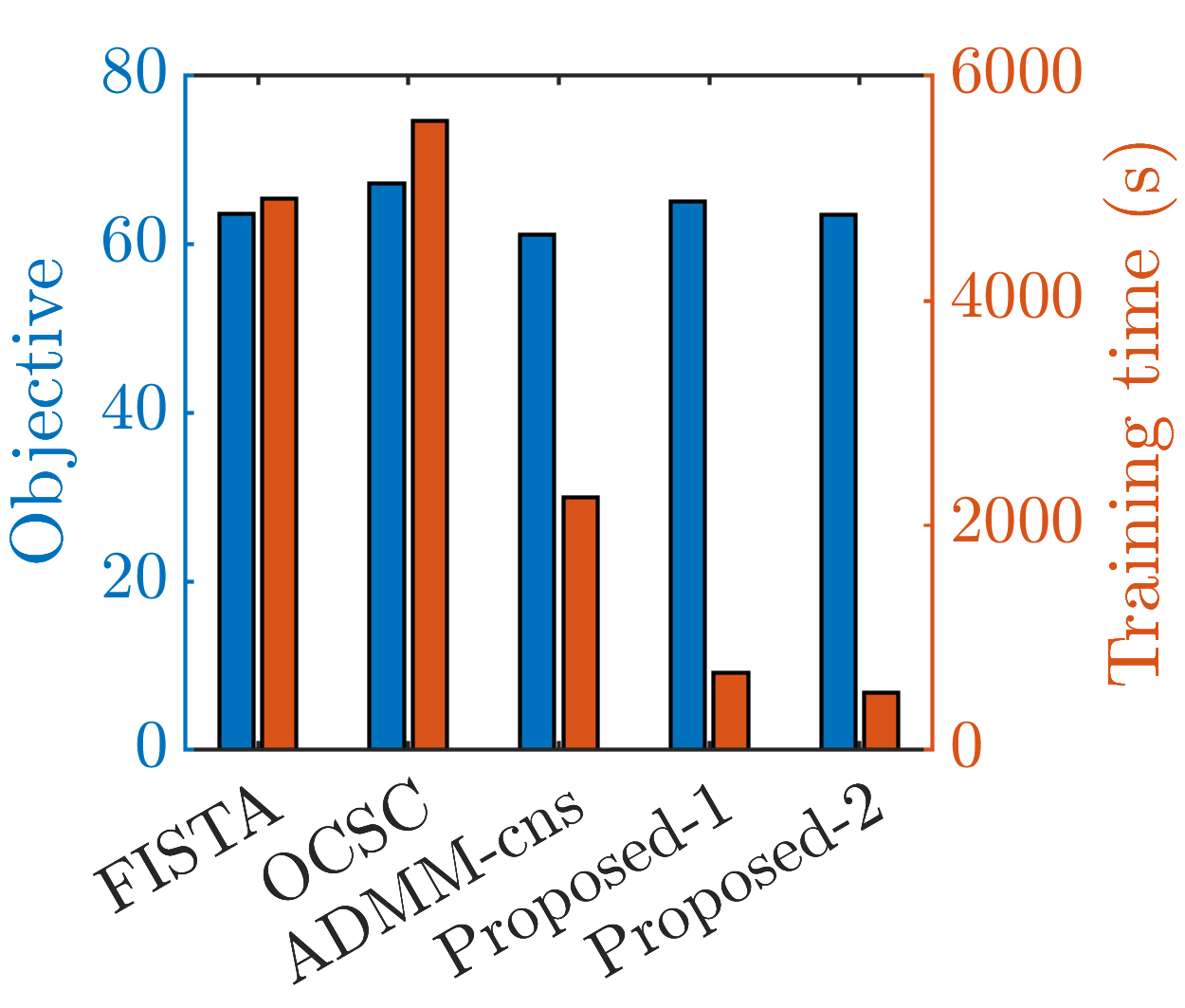}}\\
{\includegraphics[width=.65\linewidth]{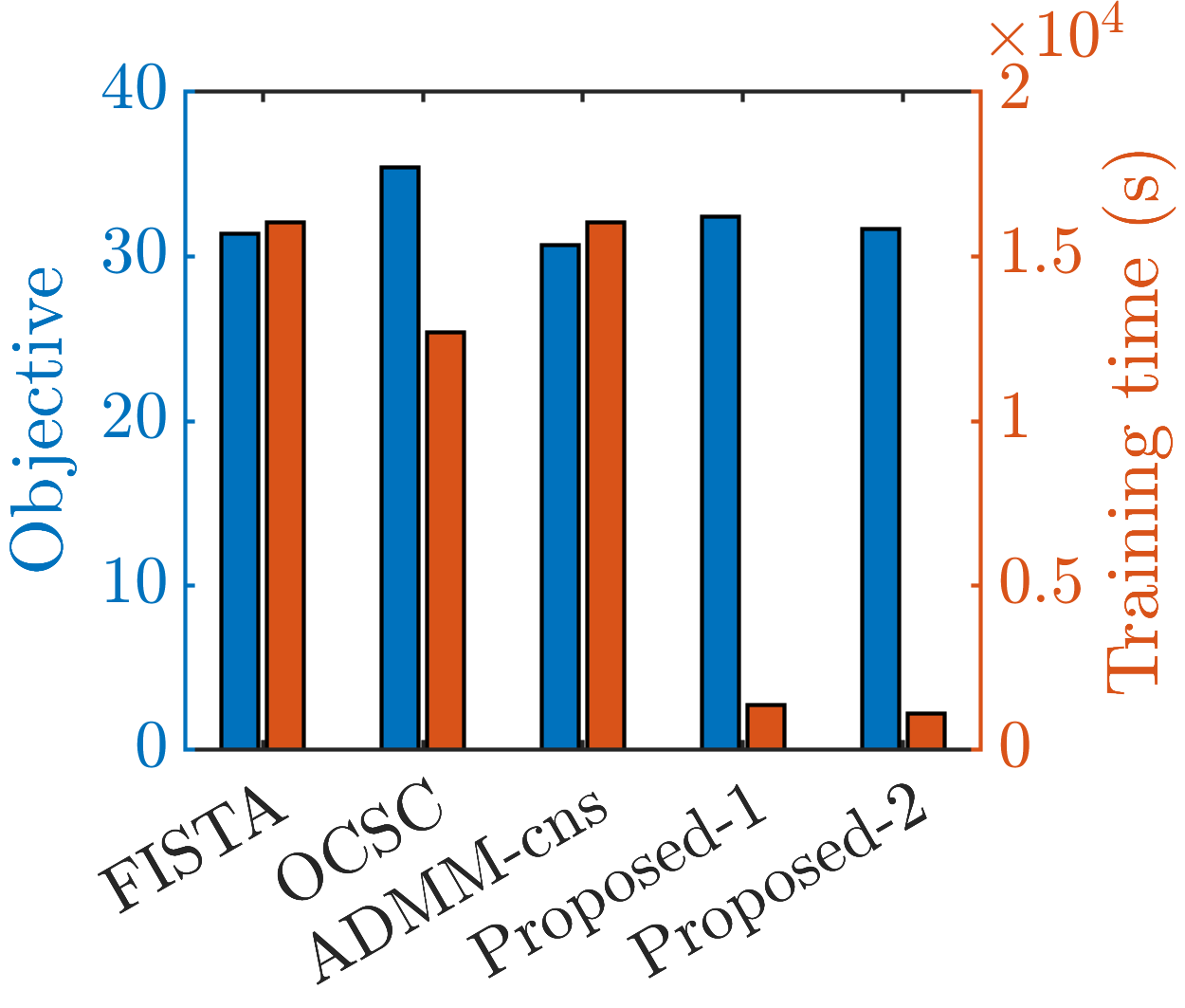}}
\caption{Comparison of test objective values and training times obtained using all methods compared for datasets \emph{SIPI} (top) and \emph{Flickr} (bottom).}
\label{fig: bars flicker and SIPI}
\end{figure}

Figs.~\ref{fig: SIPI data} and \ref{fig: Flickr data} depict $10$ images randomly selected from the \emph{SIPI} and \emph{Flickr} datasets, respectively. The experiments for \emph{SIPI} dataset are carried out using a dictionary size of $K=80$. A dictionary size of $K=100$ is used for the experiments based on \emph{Flickr} dataset. The average test objective values and the training times obtained using all methods tested for these two datasets are reported in~\Cref{tab: sipi,tab: flickr}, and displayed in bar charts in Fig.~\ref{fig: bars flicker and SIPI}.


\begin{table}[htb]
\caption{Average test objective values and training times obtained using the methods compared for dataset \emph{SIPI}.}
\centering
\setlength{\tabcolsep}{10pt}
\renewcommand{\arraystretch}{1}
\begin{tabular}{|c|c|c|}
\hline & Objective & Training Time (s) \\ \hline
Initial dictionary         & $103.0952$   & -                 \\ \hline
FISTA~\cite{liu2018first}  & $63.6088$   & $4904$        \\ \hline
OCSC~\cite{Wang2018}       & $67.2540$   & $5598$        \\ \hline
Proposed-1                 & $65.0985$   & $685$         \\ \hline
Proposed-2                 & $63.4867$   & $513$         \\ \hline
ADMM-cns (batch)~\cite{veshki2021efficient} & $61.1713$   & $2248$   \\ \hline
\end{tabular}\label{tab: sipi}
\end{table}

\begin{table}[htb]
\caption{Average test objective values and training times obtained using the methods compared for dataset \emph{Flickr}.}
\centering
\setlength{\tabcolsep}{10pt}
\renewcommand{\arraystretch}{1}
\begin{tabular}{|c|c|c|}
\hline & Objective & Training Time (s) \\ \hline
Initial dictionary         & $51.6432$   & -                 \\ \hline
FISTA~\cite{liu2018first}  & $31.3904$   & $16032$        \\ \hline
OCSC~\cite{Wang2018}       & $35.4325$   & $12689$        \\ \hline
Proposed-1                 & $32.4064$   & $1362$         \\ \hline
Proposed-2                 & $31.6799$   & $1102$         \\ \hline
ADMM-cns (batch)~\cite{veshki2021efficient} & $30.6657$   & $16049$   \\ \hline
\end{tabular}\label{tab: flickr}
\end{table}

As can be seen in~\Cref{tab: sipi,tab: flickr}, the ADMM-cns method achieves the lowest test objective values. However, its advantage over the OCDL methods is not as noticeable as in the case of experiments on small datasets \emph{Fruit} and \emph{City}. Specifically, in the experiments on the larger dataset \emph{Flickr}, ADMM-cns performs only slightly better than FISTA and proposed-2, while requiring the longest training time. Among the OCDL methods, FISTA results in the smallest test objective in the experiments on \emph{Flickr}, although it takes the longest training time. The proposed methods result in comparable test objective values to other OCDL methods while substantially shortening the training time. In particular, Algorithm-2 has the smallest objective among all OCDL algorithms for the \emph{SIPI} dataset.

\begin{figure}[htb]
\centering{
\subfloat[ADMM-cns]{\includegraphics[width=.3\linewidth]{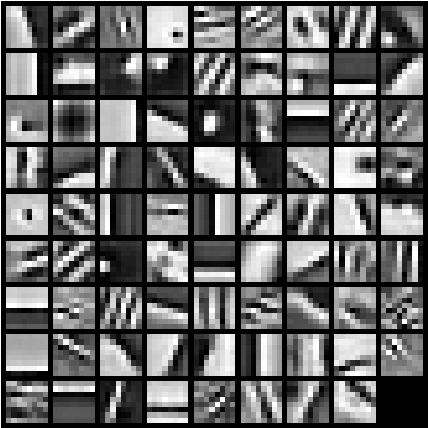}}\,
\subfloat[FISTA]{\includegraphics[width=.3\linewidth]{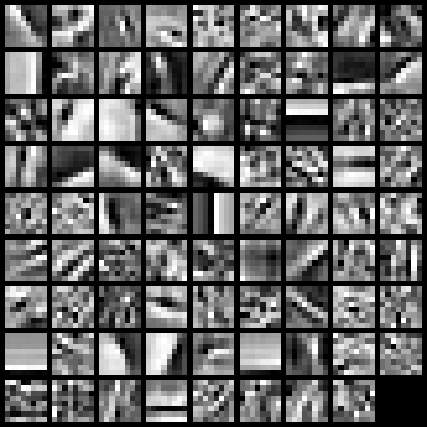}}\,
\subfloat[OCSC]{\includegraphics[width=.3\linewidth]{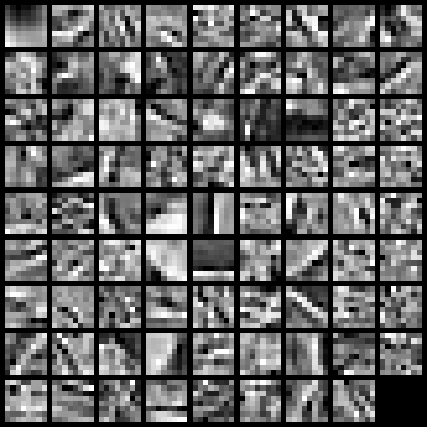}}\\\vspace{0.5mm}
\subfloat[Proposed-1]{\includegraphics[width=.3\linewidth]{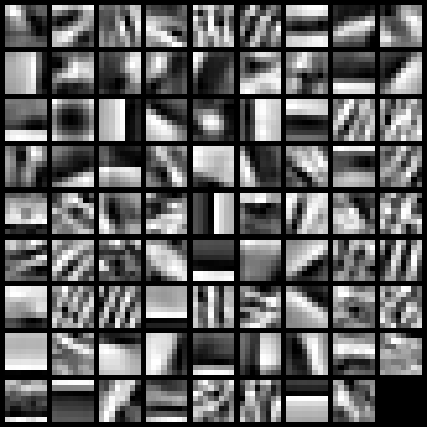}}\,
\subfloat[Proposed-2]{\includegraphics[width=.3\linewidth]{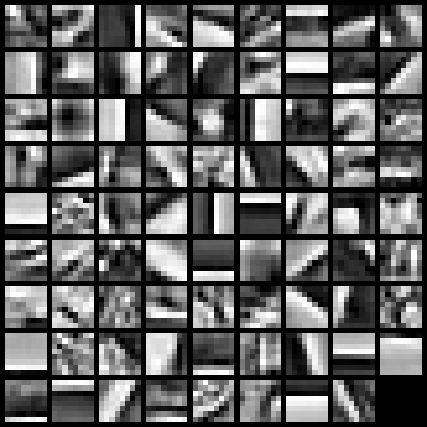}}}
\caption{Dictionaries learned ($K=80$) using the methods compared for dataset \emph{SIPI}.}
\label{fig: SIPI dicts}
\end{figure}

\begin{figure}[htb]
\centering{
\subfloat[ADMM-cns]{\includegraphics[width=.3\linewidth]{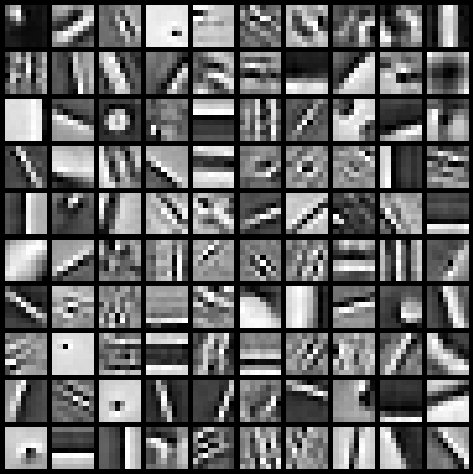}}\,
\subfloat[FISTA]{\includegraphics[width=.3\linewidth]{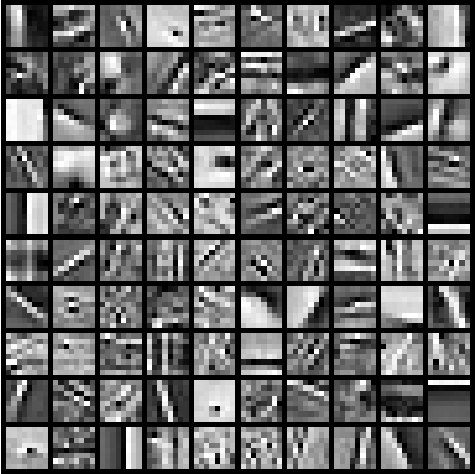}}\,
\subfloat[OCSC]{\includegraphics[width=.3\linewidth]{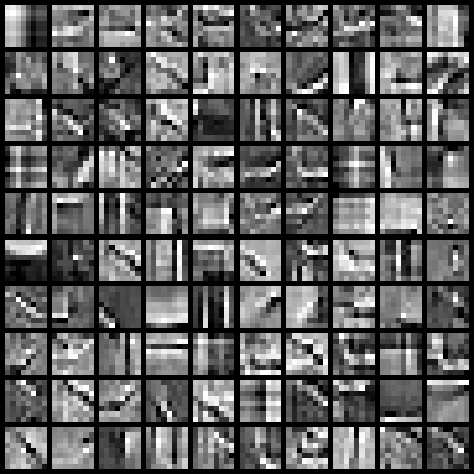}}\\\vspace{0.5mm}
\subfloat[Proposed-1]{\includegraphics[width=.3\linewidth]{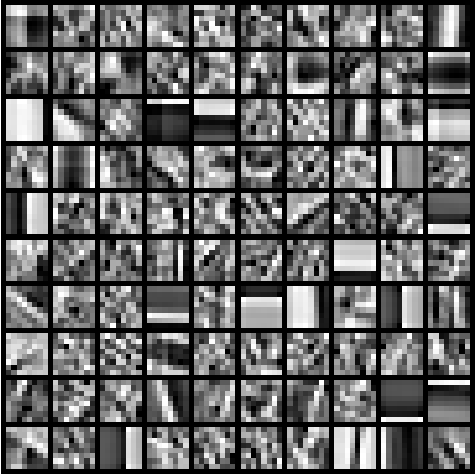}}\,
\subfloat[Proposed-2]{\includegraphics[width=.3\linewidth]{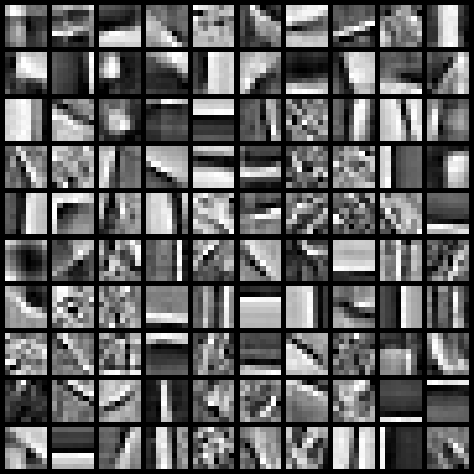}}}
\caption{Dictionaries learned ($K=100$) using the methods compared for dataset \emph{Flickr}.}
\label{fig: flickr dicts}
\end{figure}

The convolutional dictionaries learned based on datasets \emph{SIPI} and \emph{Flickr} using the methods tested are shown in Figs.~\ref{fig: SIPI dicts} and \ref{fig: flickr dicts}, respectively. As can be observed from the dictionaries displayed in Fig.~\ref{fig: SIPI dicts}, in the experiments on \emph{SIPI}, the dictionary filters learned using the proposed algorithms are less noisy and random compared to those learned using FISTA and OCSC. 
For the experiment on the \emph{Flicker} dataset, the dictionary filters learned using FISTA are crisper and sharper compared to other OCDL methods tested (FISTA also resulted in the smallest test objective for dataset \emph{Flickr}).


\subsection{Learning Large Dictionaries}

\begin{figure*}[htb]
\centering
{\includegraphics[width=.224\linewidth]{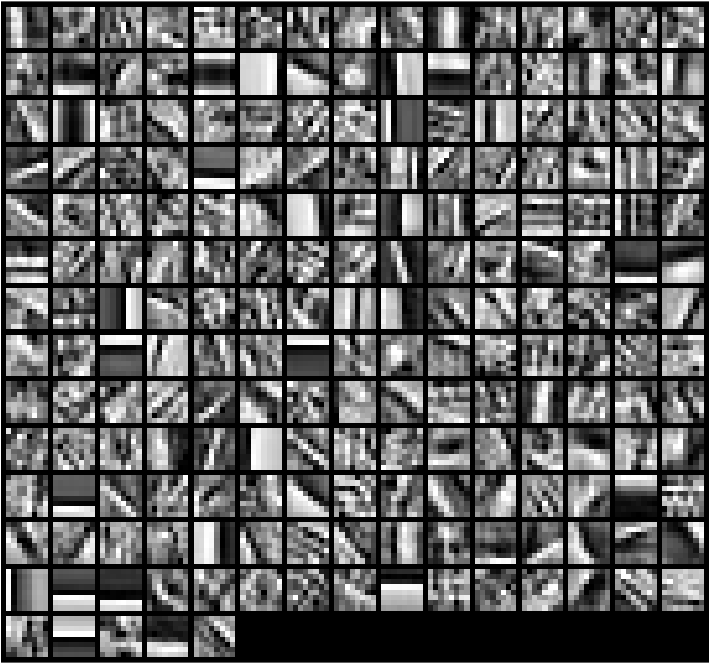}}\;
{\includegraphics[width=.264\linewidth]{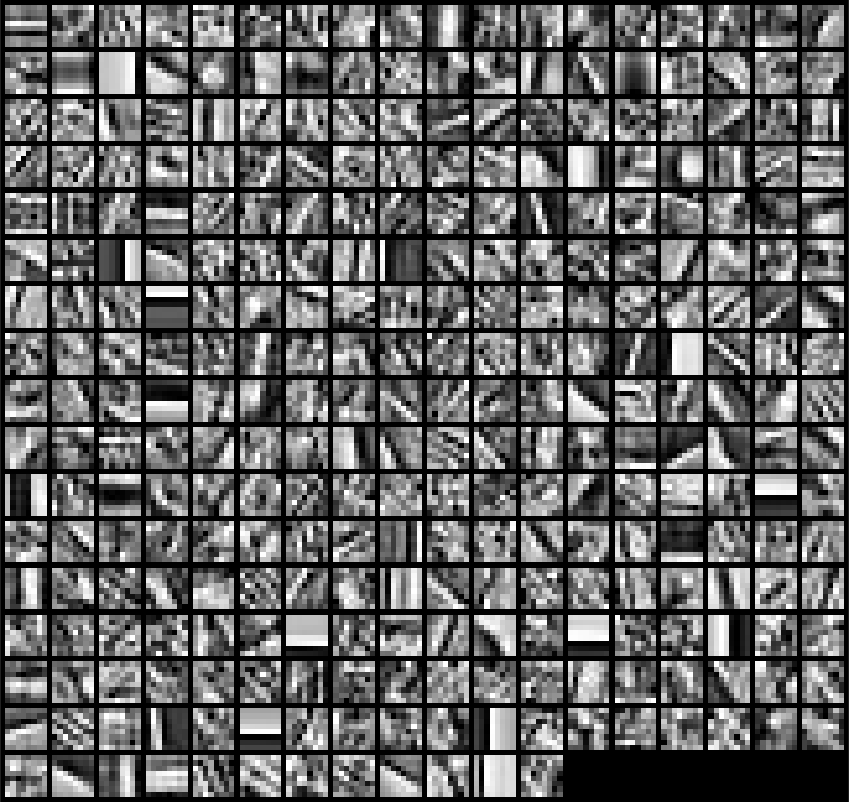}}\;
{\includegraphics[width=.296\linewidth]{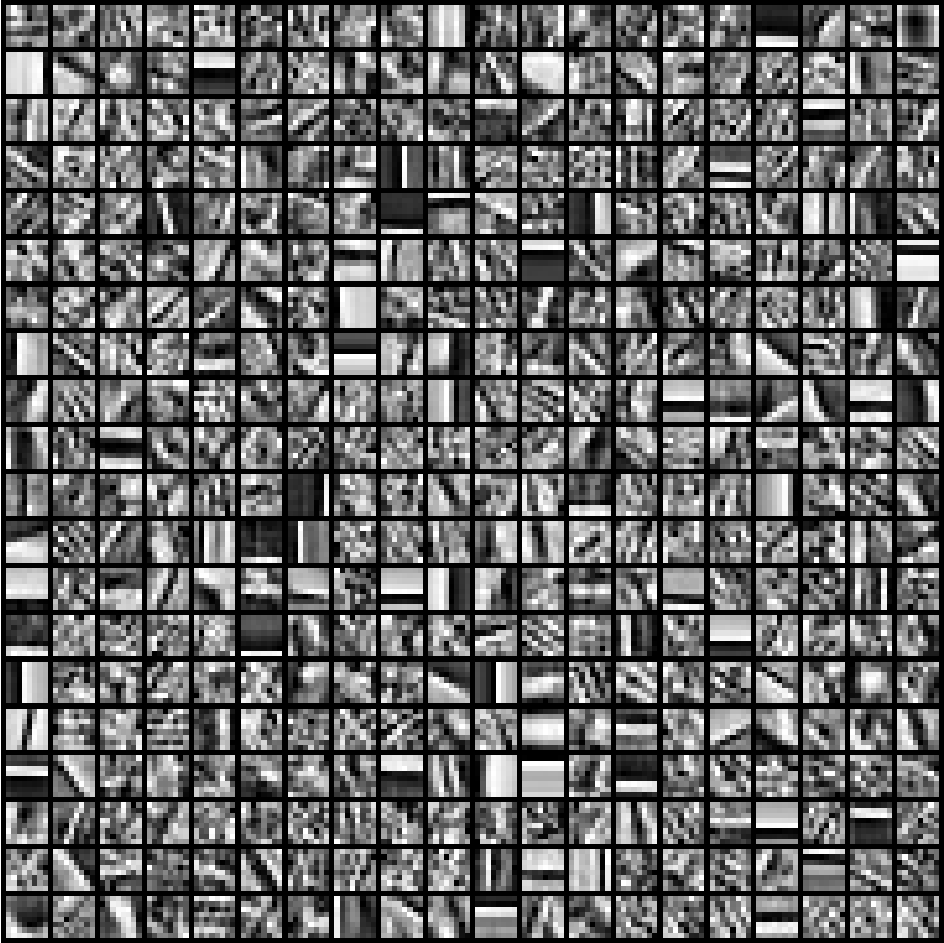}}\\\vspace{1mm}
{\includegraphics[width=.224\linewidth]{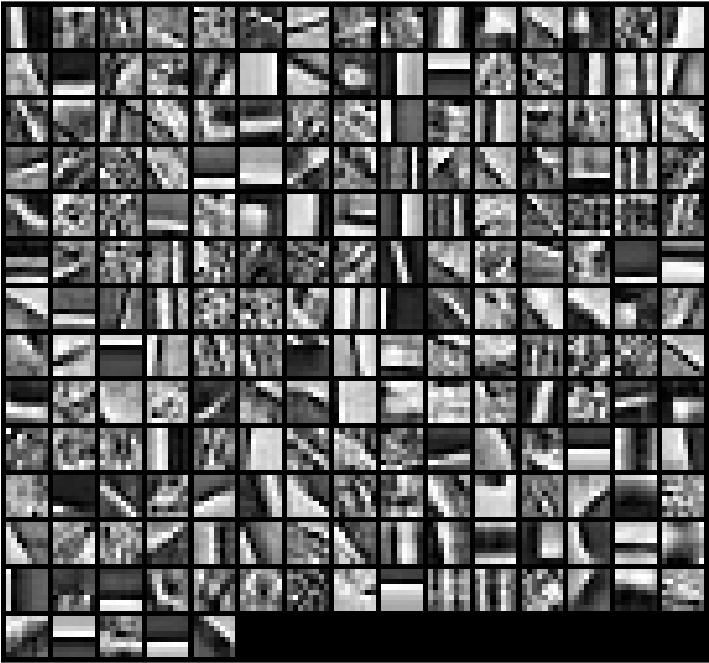}}\;
{\includegraphics[width=.264\linewidth]{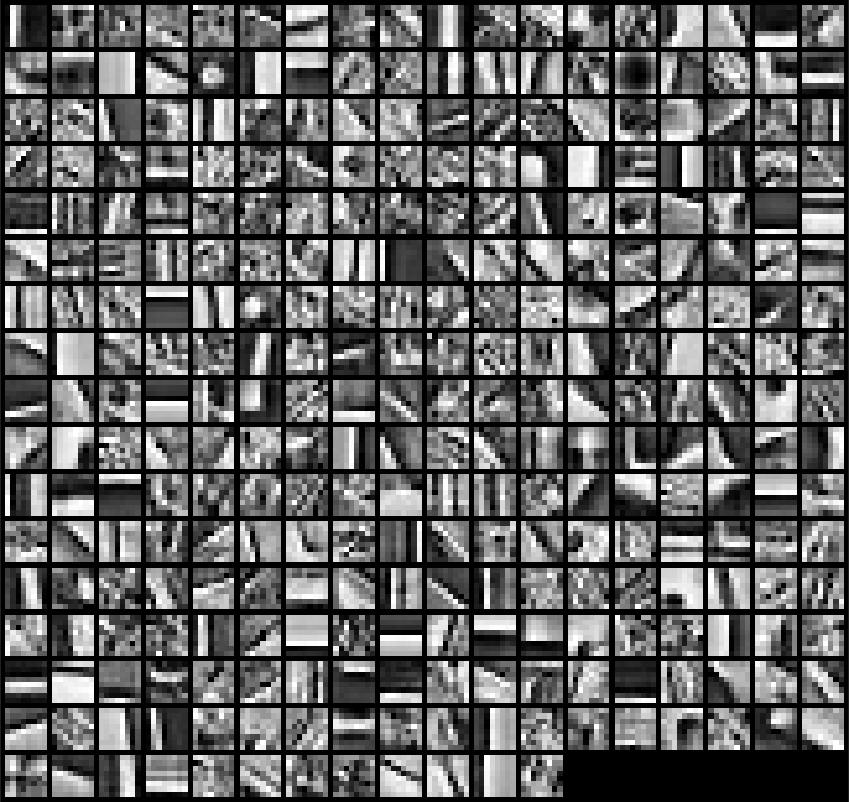}}\;
{\includegraphics[width=.296\linewidth]{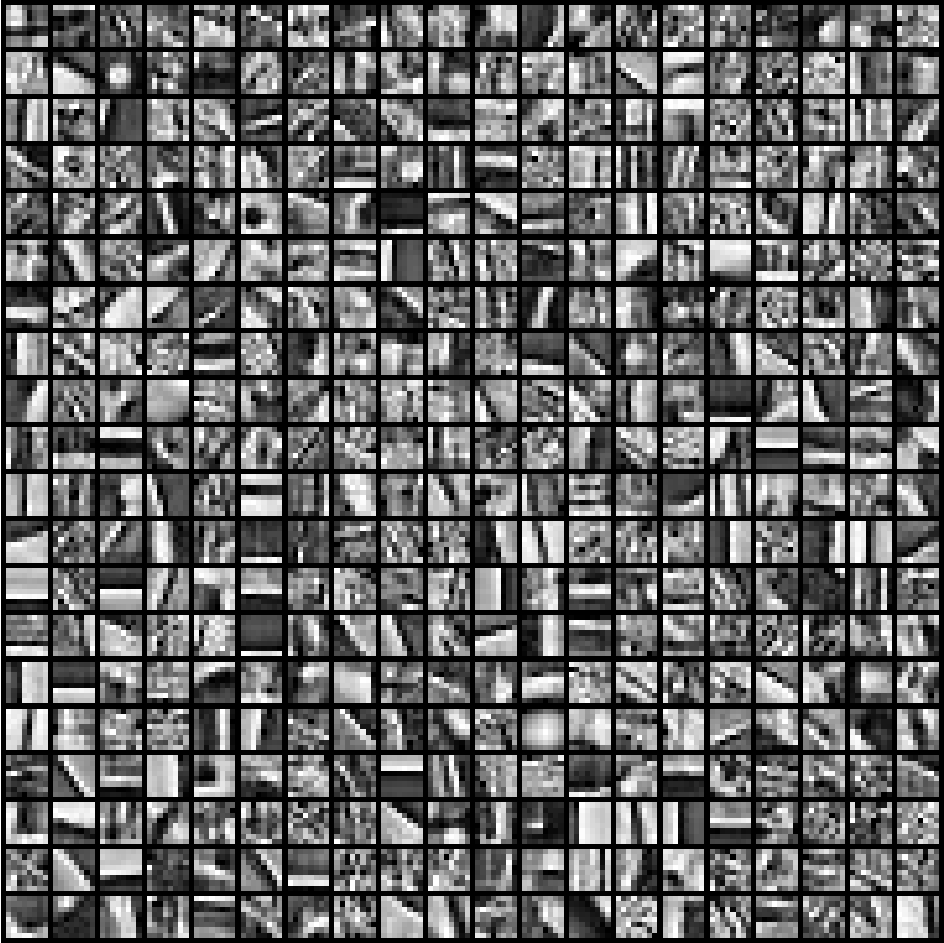}}
\caption{Large dictionaries learned using the proposed algorithms (top: proposed-1, bottom: proposed-2) for dataset \emph{Flickr} with $K=200$ (left), $K=300$ (middle), and $K=400$ (right).}
\label{fig: large dicts}
\end{figure*}

\begin{figure}[htb]
\centering
{\includegraphics[width=.75\linewidth]{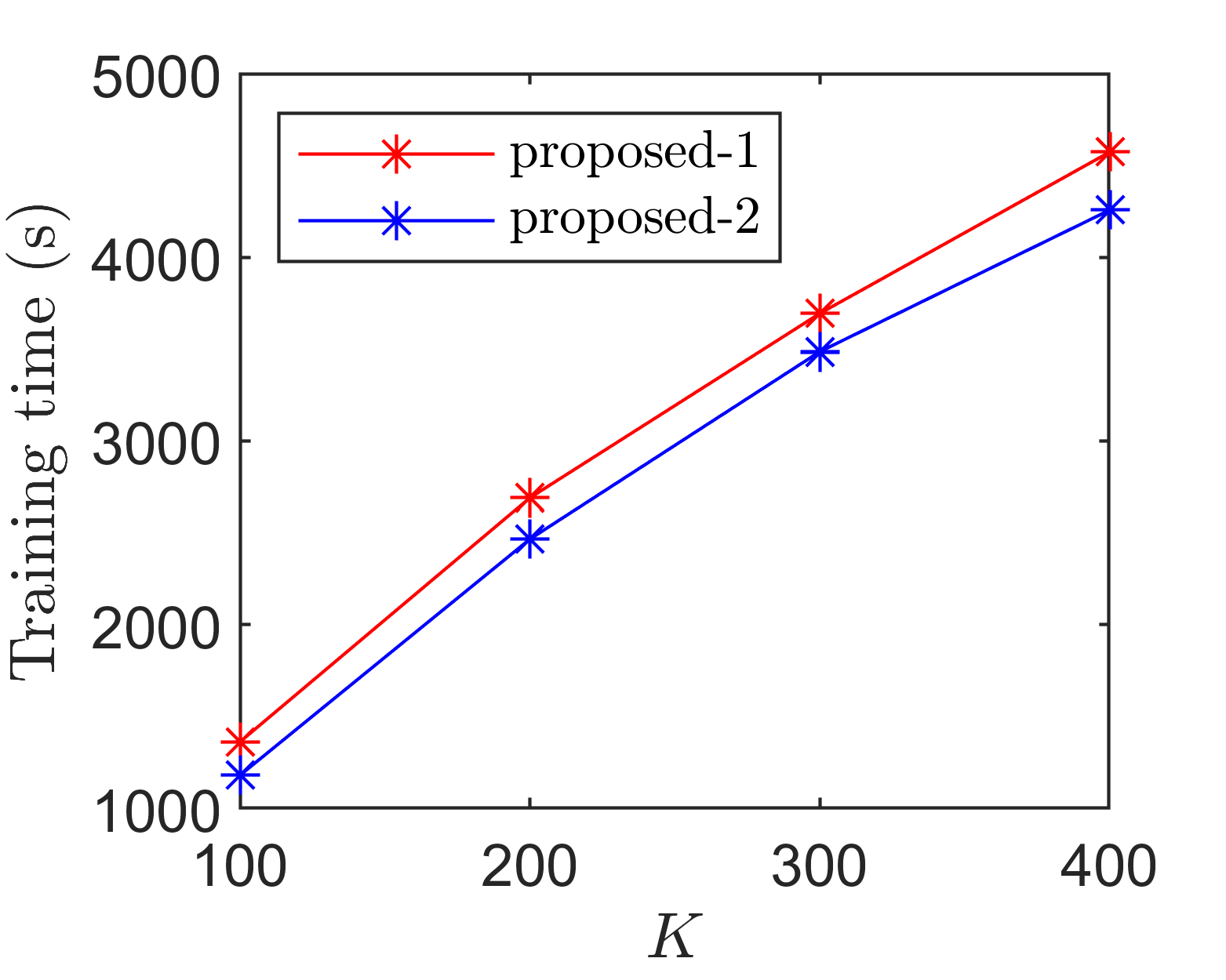}}
\caption{Comparison of training times obtained using the proposed algorithms and dataset \emph{Flickr} for learning dictionaries of different sizes.}
\label{fig: large dicts time}
\end{figure}

In this experiment, we use the proposed algorithms to learn large dictionaries of sizes $K=200$, $K=300$, and $K=400$ based on the \emph{Flickr} dataset. Learning such large dictionaries over the images of the size of those in \emph{Flickr} is not feasible using the OCDL methods, OCSC and FISTA. Indeed, in single precision, for $K=200$, only the larger history array of these methods, that is of size $K^2P$, would require \emph{more than $10$ Gigabytes memory}. The learned large dictionaries are visualized in Fig.~\ref{fig: large dicts}. It can be seen that all dictionaries learned are mostly composed of visually valid features. 
The obtained training times are reported in~\Cref{tab: large dicts} and Fig.~\ref{fig: large dicts time}. As can be seen, the longest training times obtained using the proposed methods are still significantly shorter than those resulting from using other methods tested for learning smaller dictionaries (see~\Cref{tab: flickr}, for example).

\begin{table}[htb]
\caption{Training times (seconds) obtained using the proposed methods for dataset \emph{Flickr}.}
\centering
\setlength{\tabcolsep}{10pt}
\renewcommand{\arraystretch}{1}
\begin{tabular}{|c|c|c|c|}
\hline & $K=200$ & $K=300$ & $K=400$ \\ \hline
Proposed-1                 & $2691$   & $3695$     & $4574$     \\ \hline
Proposed-2                 & $2466$   & $3487$    & $4258$      \\ \hline
\end{tabular}\label{tab: large dicts}
\end{table}

\subsection{CDL Over a Large Dataset}
\begin{figure}[htb]
\centering
{\includegraphics[width=.7\linewidth]{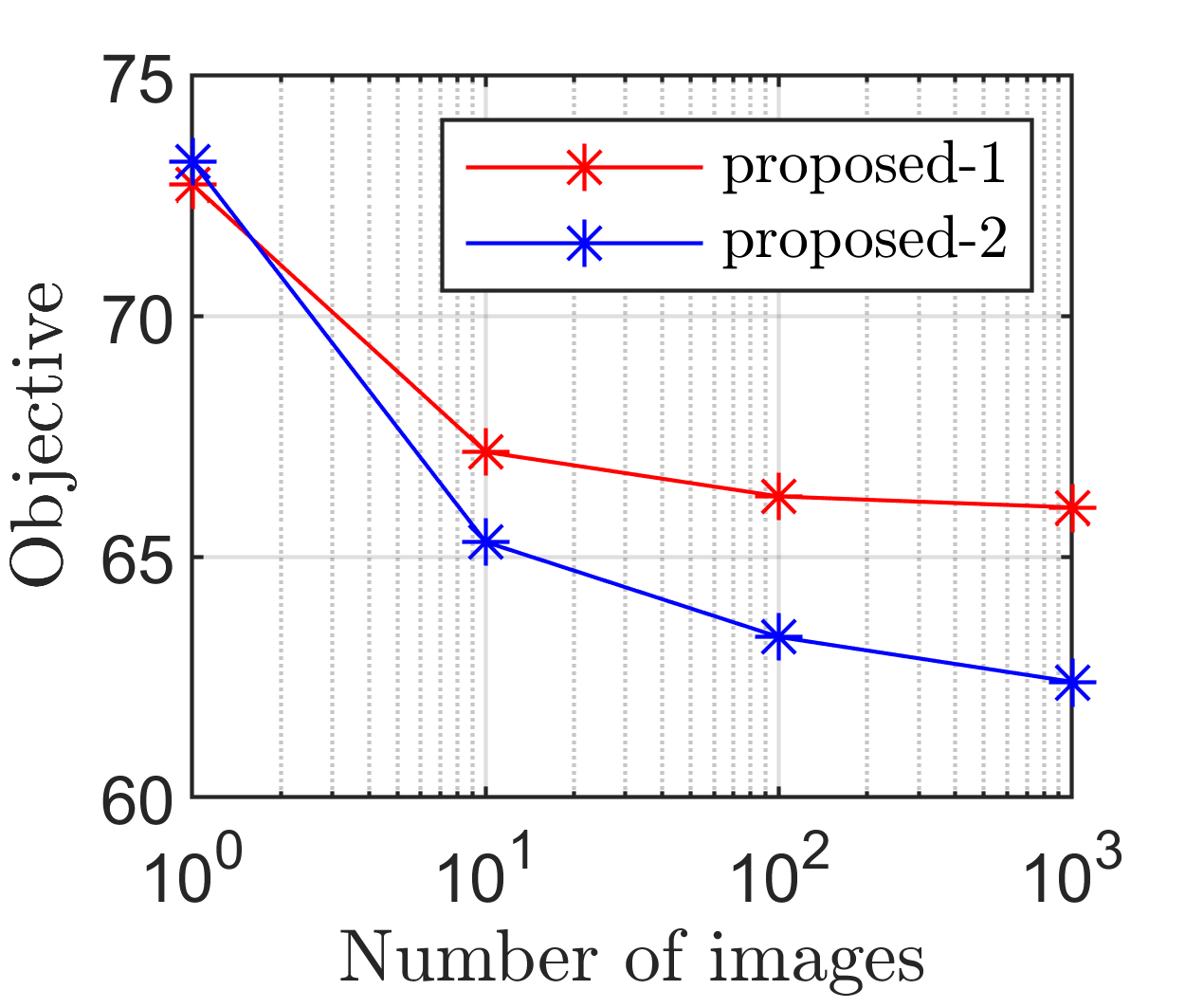}}\\
{\includegraphics[width=.7\linewidth]{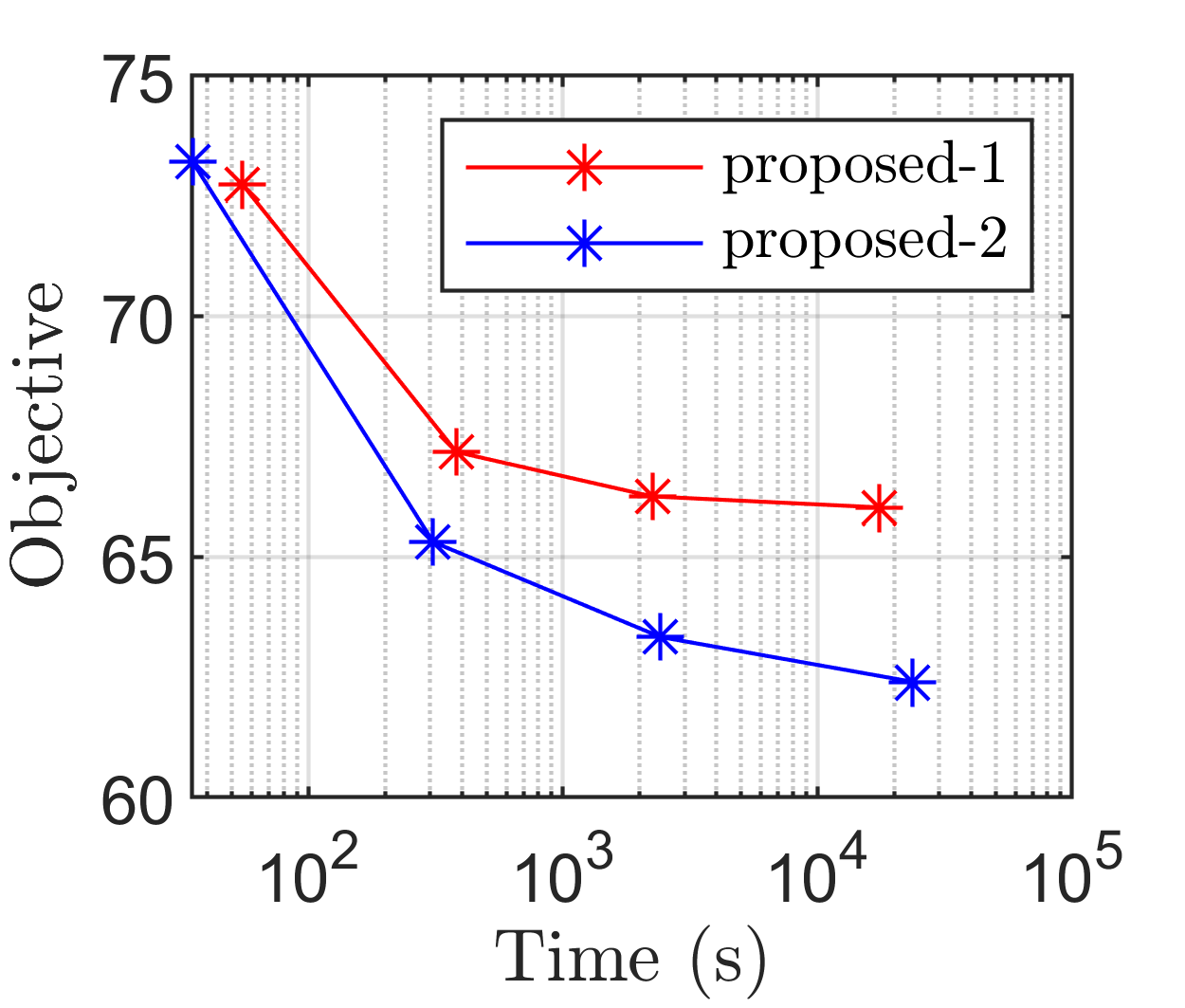}}
\caption{Results for CDL on \emph{Flickr-large} dataset using the proposed algorithms: average test objective values over the number of processed training images (top) and training time (bottom).}
\label{fig: large dataset}
\end{figure}

In this section, we demonstrate the scalability of the proposed algorithms using the \emph{Flickr-large} dataset (with $1000$ training images). Dictionaries composed of $K=100$ filters are used in this experiment. Fig.~\ref{fig: large dataset} shows the average test objective values obtained using the learned dictionaries after processing $1$, $10$, $100$, and $1000$ images. The results show that both proposed algorithms are applicable to large training datasets. However, Algorithm-2 leads to considerably lower objective values.  

\section{Conclusion}
\label{sec: conclusions}
An efficient approximate method for CDL has been presented. The proposed method is based on a novel formulation of the CDL problem that incorporates approximate sparse decomposition of training data samples. We have developed two computationally efficient OCDL algorithms based on ADMM to address the proposed approximate CDL problem. The proposed OCDL algorithms substantially reduce the required memory and improve the computational complexities of the state-of-the-art CDL algorithms. Extensive experimental evaluations using multiple image datasets have demonstrated the effectiveness of the proposed OCDL algorithms.


\bibliographystyle{IEEEtran}
\bibliography{ref}

\end{document}